\pdfoutput=1
\documentclass[11pt]{article}

\usepackage[]{EACL2023}

\usepackage{times}
\usepackage{latexsym}

\usepackage{microtype}
\usepackage{graphicx}
\usepackage{subcaption}
\usepackage{booktabs}

\usepackage{multicol}

\usepackage{amsmath}
\usepackage{amssymb}
\usepackage{mathtools}
\usepackage{amsthm}
\usepackage{algorithm, algpseudocode}
\usepackage{multirow, mathtools}
\usepackage{tabulary}
\usepackage{wrapfig, makecell}
\usepackage{enumitem}
\usepackage{pifont}
\theoremstyle{plain}
\newtheorem{theorem}{Theorem}[section]

\theoremstyle{definition}
\newtheorem{definition}[theorem]{Definition}
\newtheorem{assumption}[theorem]{Assumption}

\newcommand{\vecc}{\mathbf{v}_{\theta,ij}}
\newcommand{\hess}{H_{\ell,j}}
\newcommand{\grad}{D_{\ell,j}}

\usepackage[T1]{fontenc}
\usepackage[utf8]{inputenc}

\usepackage{microtype}

\usepackage{inconsolata}

\title{Revisiting Intermediate Layer Distillation \\ for Compressing Language Models: An Overfitting Perspective}

\author{Jongwoo Ko\textsuperscript{\rm 1} \\ KAIST AI   \And 
        Seungjoon Park\textsuperscript{\rm 1} \\ KAIST AI   \And 
        Minchan Jeong\textsuperscript{\rm 1} \\ KAIST AI  
        \AND
        Sukjin Hong\textsuperscript{\rm 2} \\ KT   \And 
        Euijai Ahn\textsuperscript{\rm 2} \\ KT   \And 
        Du-Seong Chang\textsuperscript{\rm 2} \\ KT   \And
        Se-Young Yun\textsuperscript{\rm 1} \\ KAIST AI
        \AND
        {\normalfont \textsuperscript{\rm 1}\texttt{\{jongwoo.ko, sjoon.park, mcjeong, yunseyoung\}@kaist.ac.kr}} \\
        \textsuperscript{\rm 2}\texttt{\{sukjin.hong, euijai.ahn, dschang\}@kt.com}
        \vspace{10cm}
        }
\begin{document}
\maketitle

\begin{abstract}
Knowledge distillation\,(KD) is a highly promising method for mitigating the computational problems of pre-trained language models\,(PLMs).
Among various KD approaches, Intermediate Layer Distillation\,(ILD) has been a \textit{de facto standard} KD method with its performance efficacy in the NLP field.
% However, we find that existing ILD methods are prone to overfitting to training datasets, whereas these methods transfer more information than the original KD.
In this paper, we find that existing ILD methods are prone to overfitting to training datasets, although these methods transfer more information than the original KD.
% We further present investigations to mitigate overfitting, including distilling the knowledge of the last Transformer layer and conducting ILD on supplementary tasks.
Next, we present the simple observations to mitigate the overfitting of ILD: distilling only the last Transformer layer and conducting ILD on supplementary tasks.
Based on our two findings, we propose a simple yet effective consistency-regularized ILD\,(CR-ILD), which prevents the student model from overfitting the training dataset.
% Substantial experiments on distilling BERT on the GLUE benchmark and several synthetic datasets demonstrate that our ILD framework outperforms other KD techniques.
Substantial experiments on distilling BERT on the GLUE benchmark and several synthetic datasets demonstrate that our proposed ILD method outperforms other KD techniques. Our code is available at \url{https://github.com/jongwooko/CR-ILD}.
\end{abstract}
\section{Introduction}\label{sec:intro}
Recent advances in NLP have shown that using PLMs such as BERT\,\cite{devlin2018bert} and RoBERTa\,\cite{liu2019roberta} on downstream tasks is effective. 
Although these models achieve state-of-the-art performances in various domains, the promising results of PLMs require numerous computation and memory costs. 
Deploying such large models on resource-constrained devices such as mobile and wearable devices is impractical. 
It is thus crucial to train computationally efficient small-sized networks with similar performance to that of large models.

KD is promising model compression technique where knowledge is transferred from a large and high-performing model (teacher) to a smaller model (student).
KD has been shown to be reliable in reducing the number of parameters and computations while achieving competitive results on downstream tasks.
Recently, KD has attracted more attention in the NLP field, especially due to large PLMs. 
However, it is clear that the original KD\,\cite{hinton2015distilling} is not performing well in terms of maintaining the performance of compressed PLMs and that it needs to have additional auxiliary training objectives\,\cite{sun2019patient, jiao2019tinybert}.
\begin{figure*}[t]
    \hspace*{\fill}
    \begin{subfigure}[b]{0.30\textwidth} % 0.4 % 0.23
    \centering
    \includegraphics[width=\linewidth]{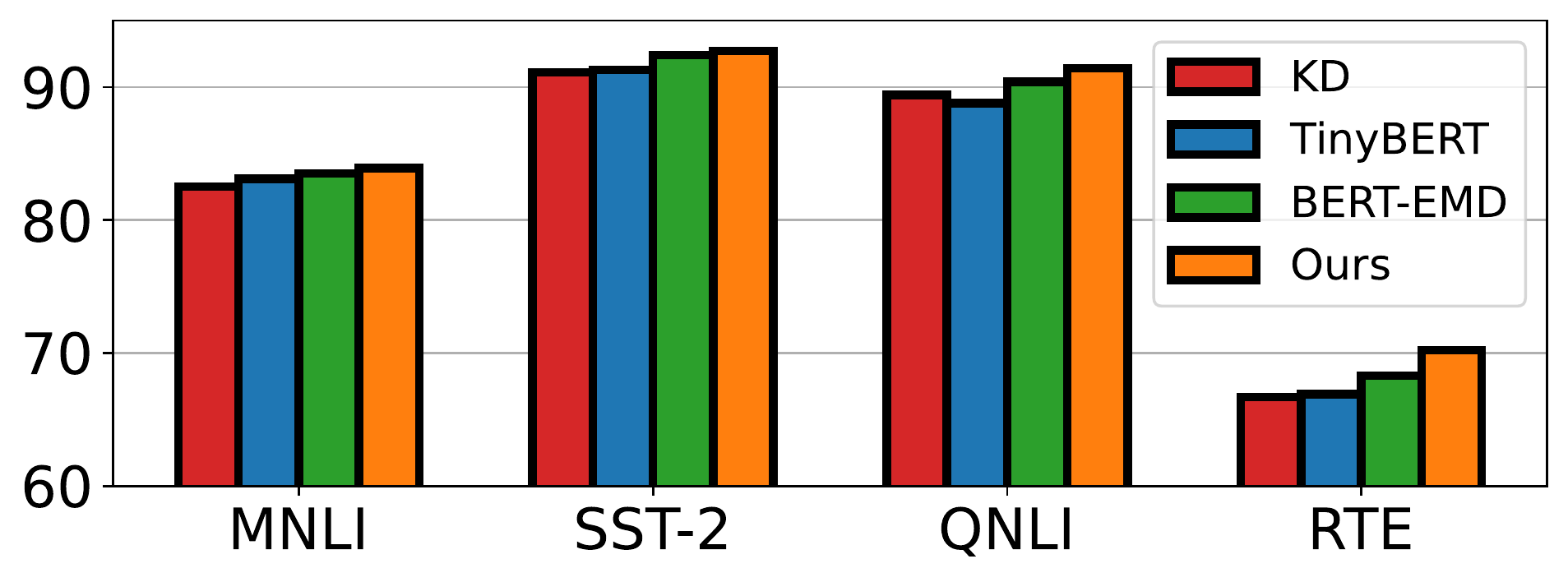}
    \caption{Standard}
    \end{subfigure}
    \hfill
    \begin{subfigure}[b]{0.30\textwidth} % 0.4 % 0.23
    \centering
    \includegraphics[width=\linewidth]{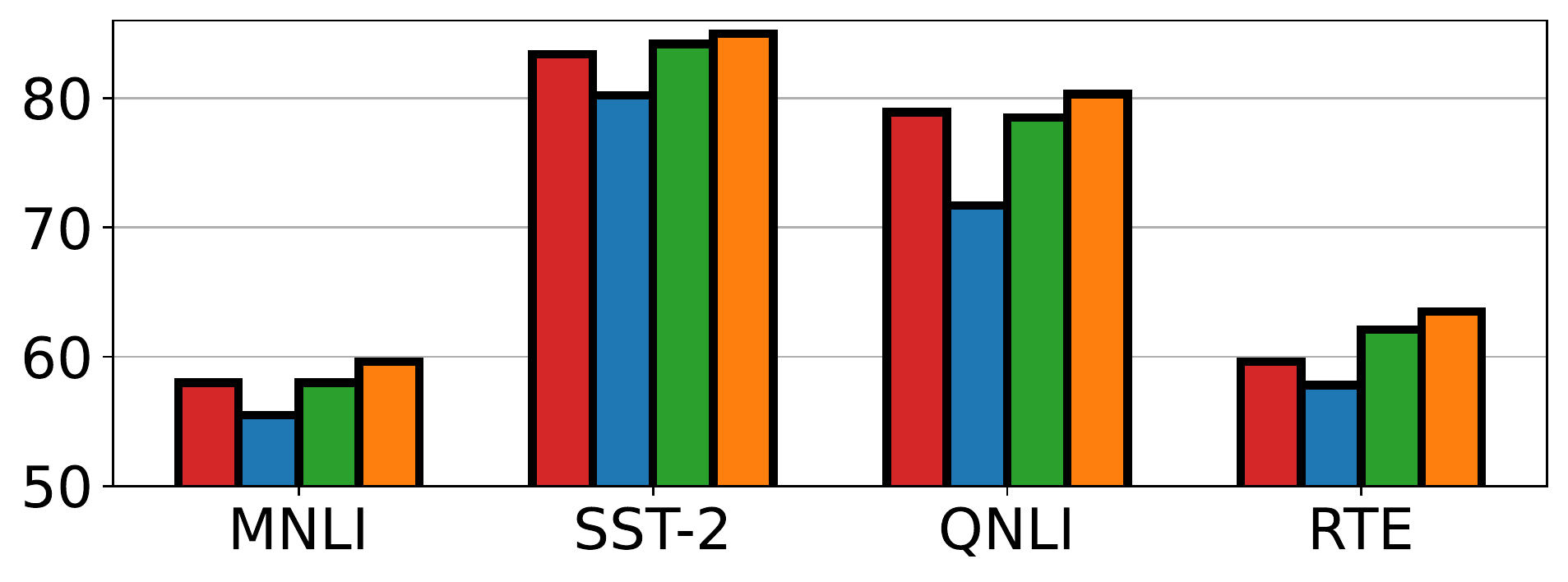}
    \caption{Few-samples}
    \end{subfigure}
    \hfill
    \begin{subfigure}[b]{0.30\textwidth} % 0.4 % 0.23
    \centering
    \includegraphics[width=\linewidth]{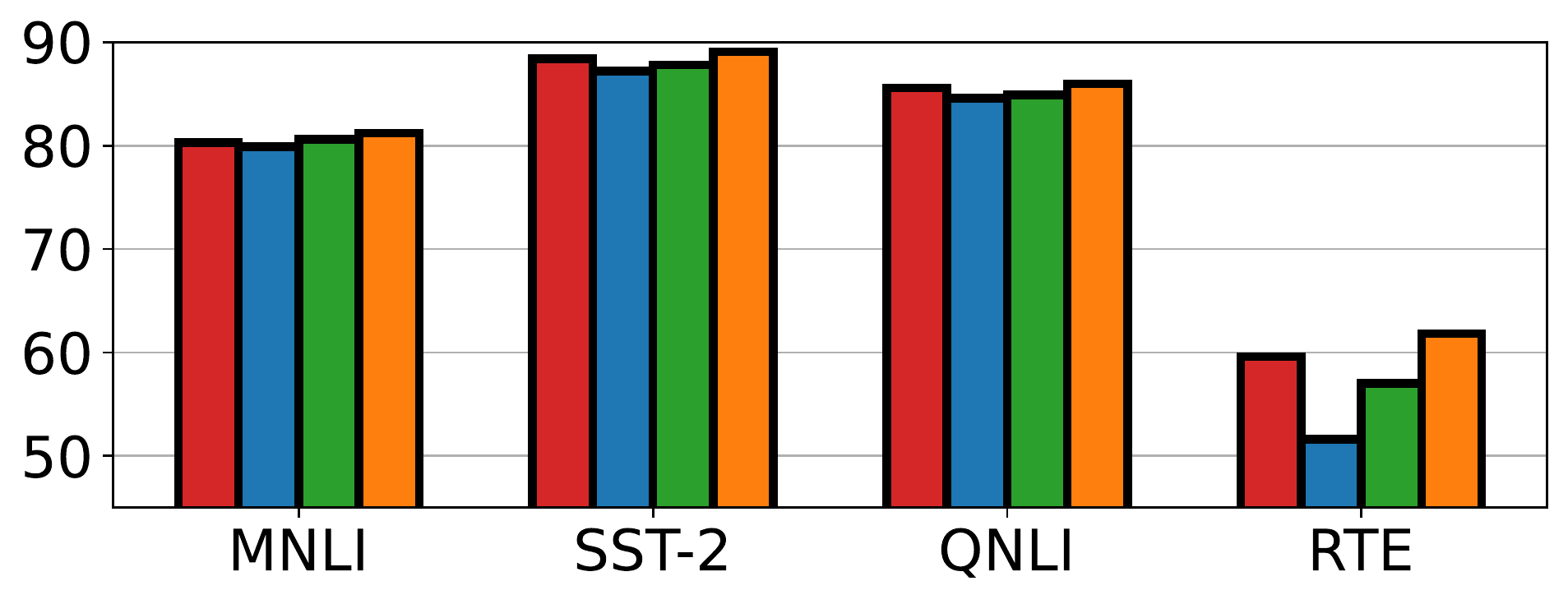}
    \caption{Label Noise}
    \end{subfigure}
    % \hfill
    % \begin{subfigure}[b]{0.24\textwidth} % 0.4 % 0.23
    % \centering
    % \includegraphics[width=\linewidth]{example-image-a}
    % \caption{Class Imbalance}
    % \end{subfigure}
    \hspace*{\fill}
    \caption{The motivation of our work. While existing ILD methods\,\cite{jiao2019tinybert, li2020bert} work well on the standard GLUE benchmark\,\cite{wang2018glue}, we observe that the existing ILD methods are problematic under few-samples training datasets or the presence of label noise. However, our proposed method shows robustly higher performance than the original KD for all datasets. We use BERT$_{\text{Small}}$\,\cite{turc2019well} as the student model and BERT$_{\text{BASE}}$\,\cite{devlin2018bert} as the teacher model. The detailed descriptions for dataset are in Appendix~\ref{app:data}.}\label{fig:motivation}
    \vspace{-5pt}
\end{figure*}

ILD methods\,\cite{jiao2019tinybert, wang2020minilm}, which encourage the student model to extract knowledge from the Transformer layers of the teacher network, have demonstrated efficacy in improving student model performance and have become a \textit{de facto standard} in KD.
Despite of success of ILD methods, many research have been proposed to design layer mapping functions\,\cite{li2020bert, wu2020skip} or new training objective\,\cite{park2021distilling} to transfer the teacher's knowledge better.
% While these ILD methods transfer more knowledge to the student model from the intermediate Transformer layers of the teacher model, we find that the use of ILD in fine-tuning may induce performance degradation in some cases.
These ILD methods transfer more knowledge to the student model from the intermediate Transformer layers of the teacher model.
However, we find that the use of ILD in fine-tuning may induce performance degradation in some cases.
As shown in \autoref{fig:motivation}, while existing ILD methods such as TinyBERT\,\cite{jiao2019tinybert} and BERT-EMD\,\cite{li2020bert} work well on standard GLUE benchmark\,\cite{wang2018glue}, we observe that these methods have performance degradation compared to original KD on ill-conditioned datasets such as those with few-samples and label noise.
Because few-sample\,\cite{zhang2020revisiting} or heterogeneous datasets\,\cite{jin2021instance, liu-etal-2022-noise} can be easily found in real-world datasets, the existing ILD methods, which show performance reduction in \autoref{fig:motivation}, are hard to use in real-world applications.

To mitigate such performance degradation, we identify the main problem as that intermediate Transformer knowledge can incur overfitting on the training dataset of the student model. We further discover that distilling only the last Transformer layer knowledge and using supplementary tasks can alleviate the overfitting. Through our observations, we finally propose a simple yet effective method, consistency-regularized ILD (CR-ILD) with several analyses. Our main contributions are:
\begin{itemize}
    \item We design and conduct comprehensive experiments to identify that overfitting is one of the main problems for performance degradation of ILD in fine-tuning. To the best of our knowledge, this is the first study to find that existing ILD methods have overfitting issues.
    \item Based on our findings, we propose the consistency regularized ILD (CR-ILD) that a student self-regularized itself from risk of overfitting from ILD. We further provide empirical (and theoretical) analyses for our proposed method.
    % \item We experimentally demonstrate that our proposed method achieves state-of-the-art performance on both standard GLUE and modified GLUE\,(few samples and label noise), despite its simplicity.
    \item We experimentally demonstrate that our proposed method achieves state-of-the-art performance on both standard GLUE and ill-conditioned GLUE\,(few samples and label noise), despite its simplicity.
\end{itemize}

\section{Related Works}\label{sec:related} % Backgrounds
% \subsection{Related Works}
% \paragraph{Transformer-based LMs.}
\paragraph{Model Compression of LMs.}
Transformer encodes contextual information for input tokens\,\cite{vaswani2017attention}.
In recent years, from the success of Transformer, Transformer-based models such as GPT\,\cite{radford2018improving}, BERT\,\cite{devlin2018bert}, and T5\,\cite{raffel2019exploring} have become a new state of the arts, driving out recurrent or convolutional networks on various language tasks. 
However, the promising results of these models are accompanied by numerous parameters, which necessitate a high computation and memory cost for inference. 
Existing compression techniques can be categorized as low-rank matrix factorization\,\cite{mao-etal-2020-ladabert}, quantization\,\cite{bai-etal-2021-binarybert}, and KD\,\cite{sun2019patient}.

\paragraph{Knowledge Distillation for LMs.}
KD is one of the most well-known neural model compression techniques.
The goal of KD is to enable the student model with fewer parameters to achieve similar performance to that of the teacher model with a large number of parameters.
In the recent few years, a wide range of different methods have been developed that apply data augmentation\,\cite{jiao2019tinybert, liang2020mixkd}, adversarial training\,\cite{rashid2021mate}, and loss terms re-weighting\,\cite{jafari2021annealing} to reduce the performance gap between the teacher and the student.
In another line in the NLP field, ILD-based methods have exhibited higher effectiveness over original KD\,\cite{hinton2015distilling} methods for compression PLMs.
\citet{sun2019patient} proposed the BERT-PKD to transfer representations of the [CLS] token of the teacher model.
\citet{jiao2019tinybert} proposed TinyBERT, which performed Transformer distillation in both pre-training and fine-tuning.
\citet{wang2020minilm} distilled the self-attention module of the last Transformer layer of the teacher. 
\citet{li2020bert} leveraged earth mover's distance (EMD) to determine the optimal layer mapping between the teacher and student networks. 
\citet{park2021distilling} presented new KD objectives that transfer contextual knowledge via two types of relationships. 
\begin{figure*}[t]
    \hspace*{\fill}
    \begin{subfigure}[b]{0.24\textwidth} % 0.4 % 0.23
    \centering
    \includegraphics[width=\linewidth]{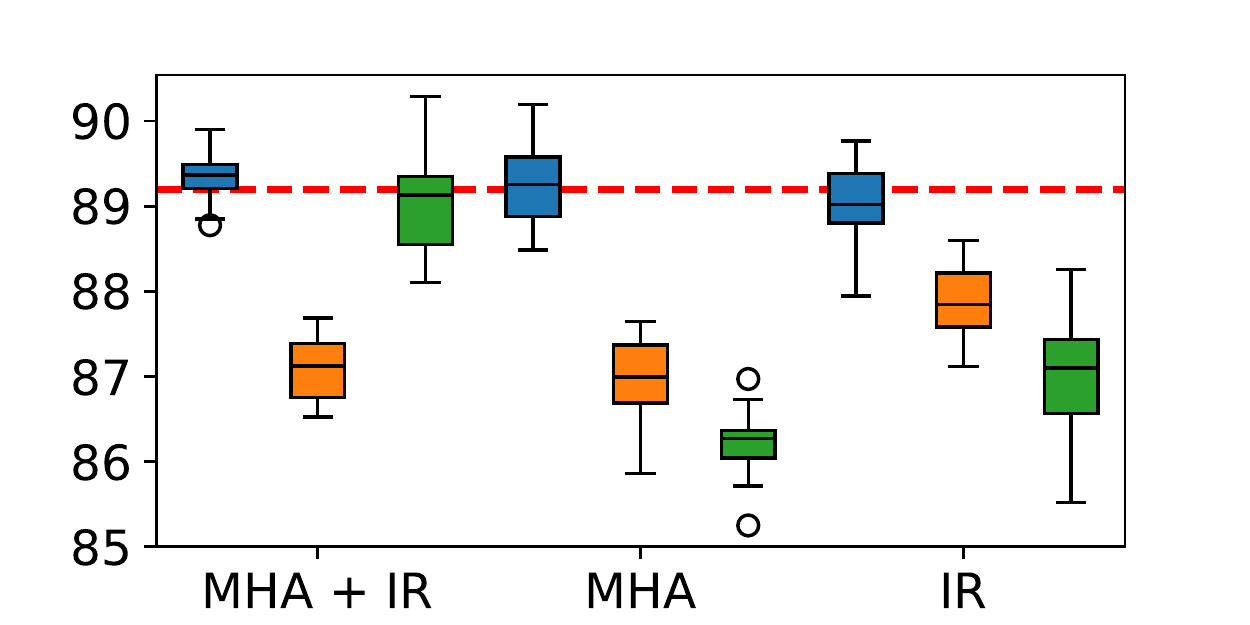}
    \caption{MRPC}
    \end{subfigure}
    \hfill
    \begin{subfigure}[b]{0.24\textwidth} % 0.4 % 0.23
    \centering
    \includegraphics[width=\linewidth]{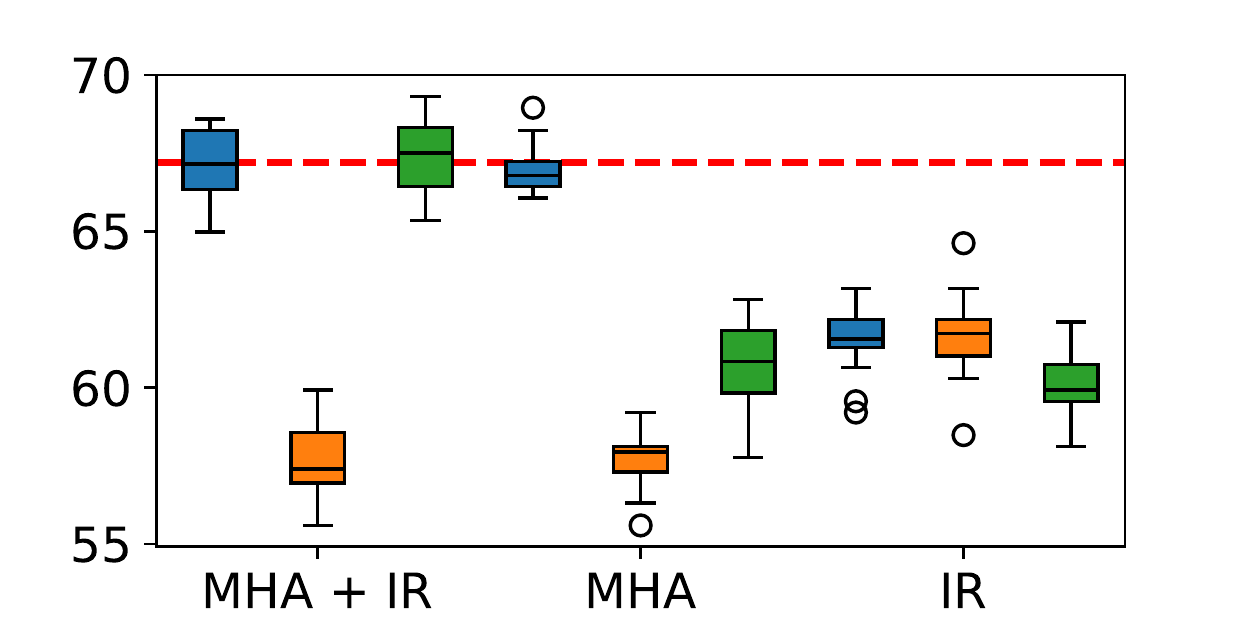}
    \caption{RTE}
    \end{subfigure}
    \hfill
    \begin{subfigure}[b]{0.24\textwidth} % 0.4 % 0.23
    \centering
    \includegraphics[width=\linewidth]{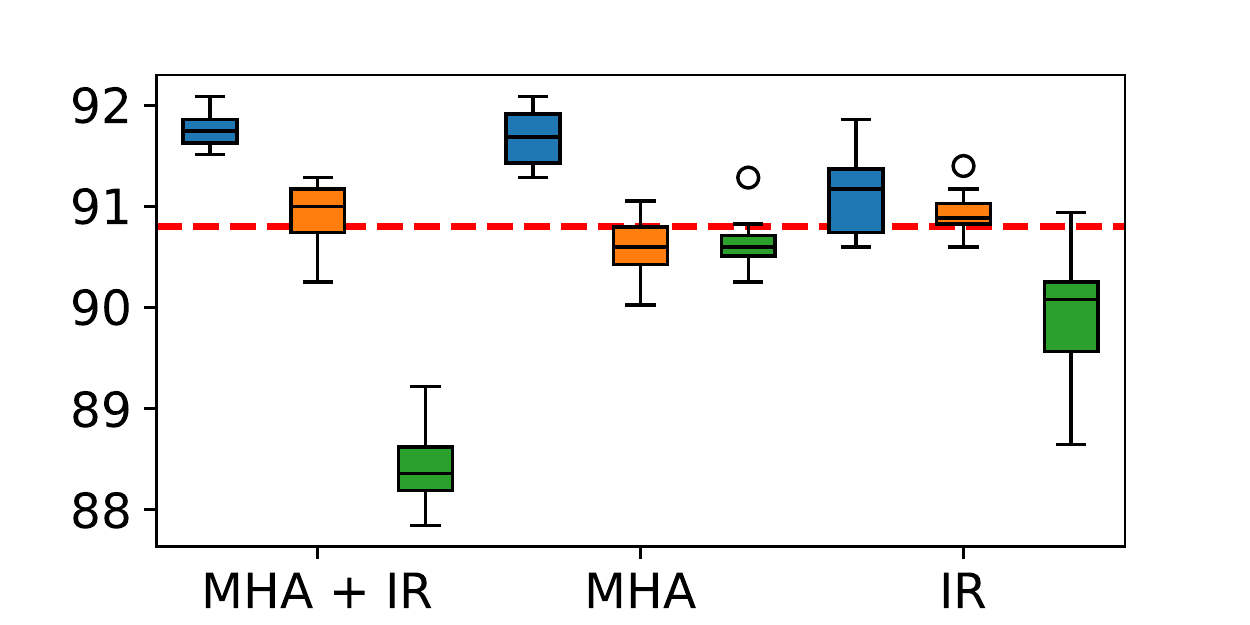}
    \caption{SST-2}
    \end{subfigure}
    \hfill
    \begin{subfigure}[b]{0.24\textwidth} % 0.4 % 0.23
    \centering
    \includegraphics[width=\linewidth]{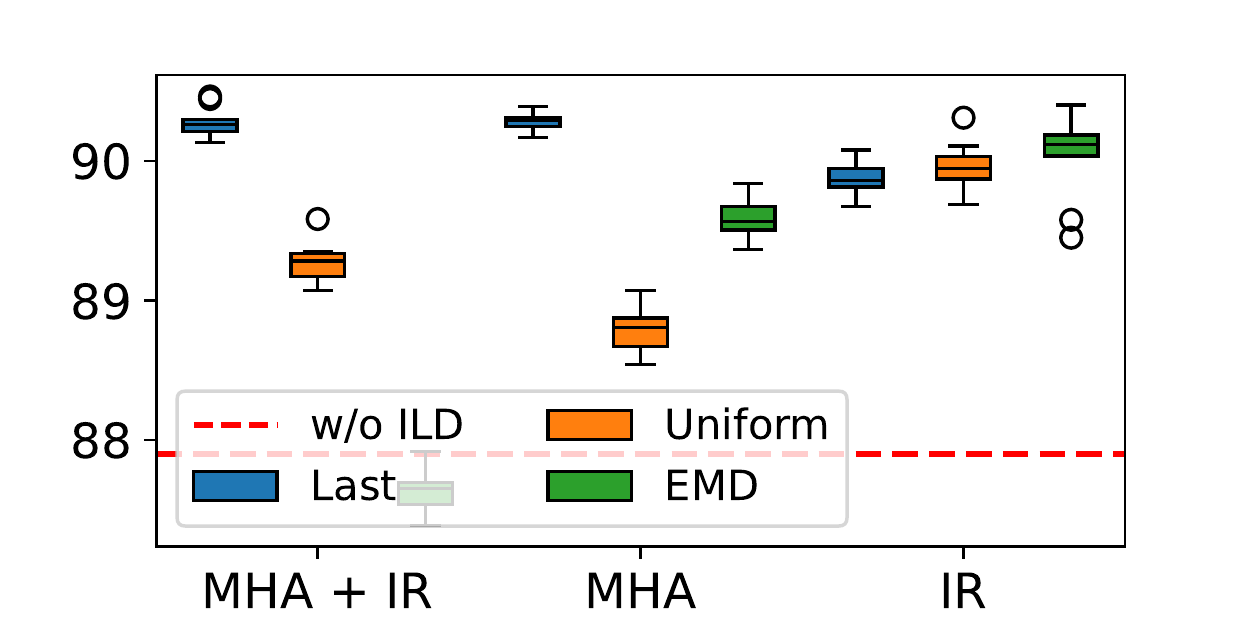}
    \caption{QNLI}
    \end{subfigure}
    \hspace*{\fill}
    \caption{Performance distribution box plot across 20 random trials and the four datasets with different distillation methods. As the student model, we apply Truncated BERT\,\cite{sun2019patient} which initialized as the bottom 6 layers from BERT$_{\text{BASE}}$. Distilling knowledge of the last Transformer layer enhances generalization and reduces the variance of fine-tuning. The red-dotted lines are baseline performances that only use prediction layer KD.}\label{fig:last_main}
\end{figure*}
% BERT Small
\begin{figure}[t]
    \hspace*{\fill}
    \begin{subfigure}[b]{0.48\linewidth} % 0.4 % 0.23
    \centering
    \includegraphics[width=\linewidth]{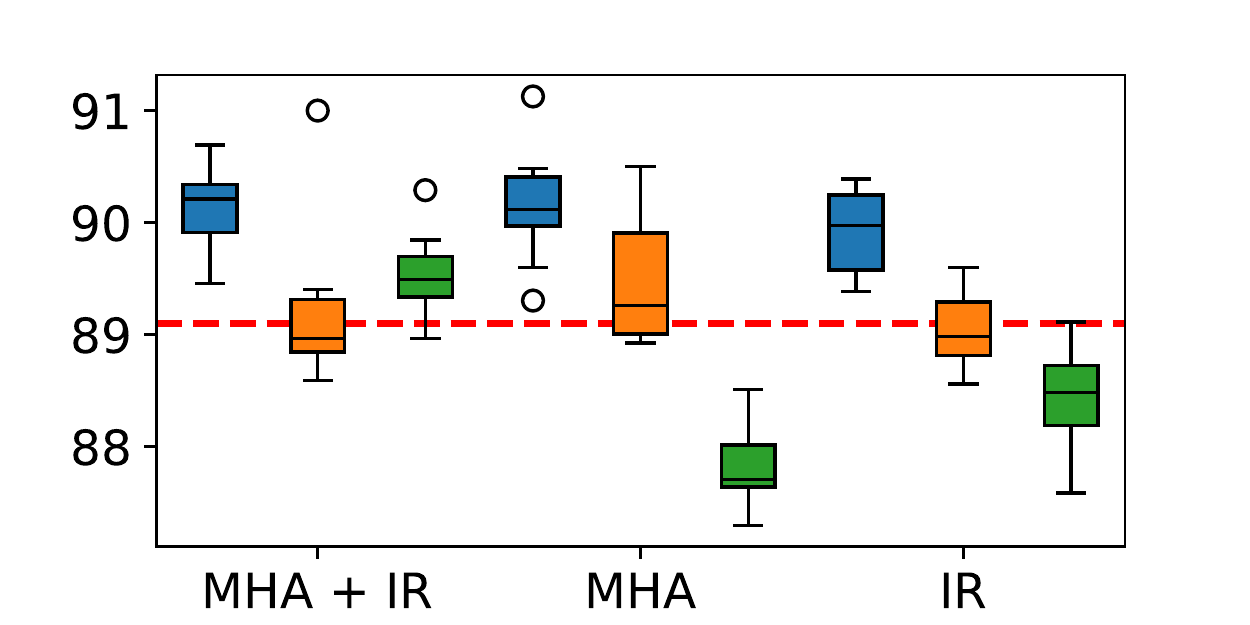}
    \caption{MRPC}
    \end{subfigure}
    \hfill
    \begin{subfigure}[b]{0.48\linewidth} % 0.4 % 0.23
    \centering
    \includegraphics[width=\linewidth]{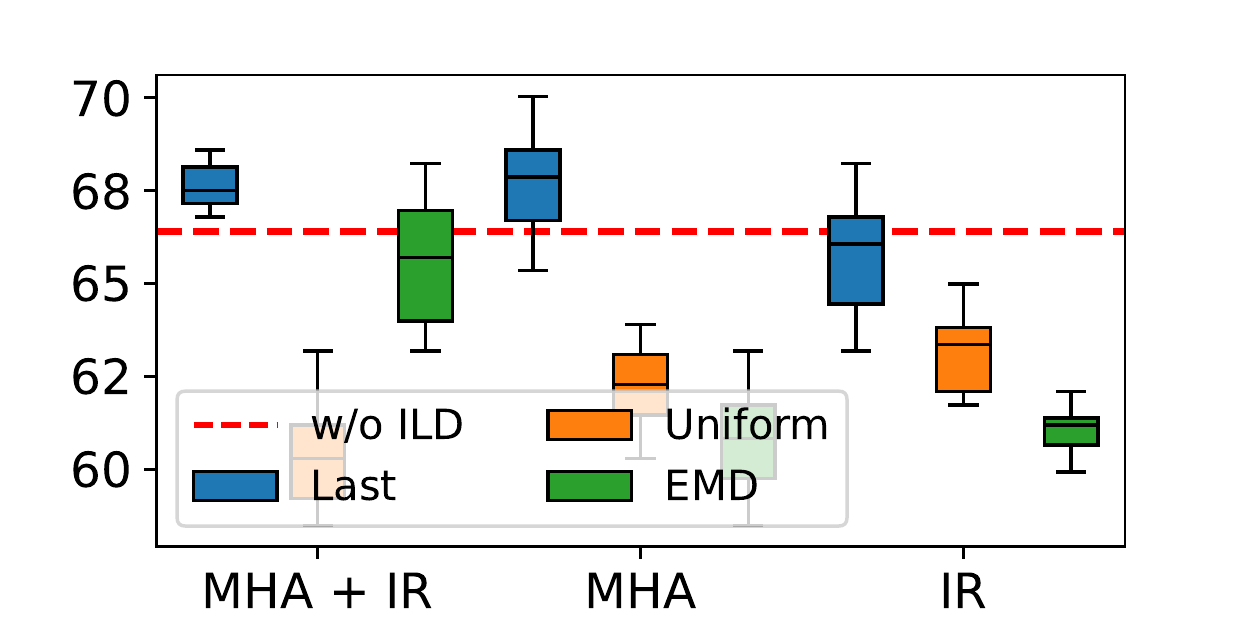}
    \caption{RTE}
    \end{subfigure}
    \hspace*{\fill}
    \caption{Performance distribution box plot across 20 random trials for MRPC and RTE when BERT$_{\text{Small}}$\,\cite{turc2019well} is used as the student model. Well-pretrained student models have more consistent performance to the choice of layer mappings.}\label{fig:last_main2}
\end{figure}
\section{Observations: Two Things Everyone Should Know to Mitigate Overfitting}\label{sec:observations} 
In this section, we identify that overfitting is the main problem for performance degradation while conducting ILD in fine-tuning.
This overfitting problem can occur even in the standard GLUE benchmark. 
Moreover, the ill-conditioned dataset, where overfitting problems can occur more easily, induces a larger performance reduction.
% Furthermore, to mitigate the risk of the overfitting problem of ILD, \textcolor{red}{we suggest two simple techniques: (1) distilling the last Transformer layer and (2) conducting ILD on supplementary tasks.}
Furthermore, we investigate that this overfitting problem is able to be reduced by (1) distilling the last Transformer layer and (2) conducting ILD on supplementary tasks.
While our suggested findings already have worked well in various domains\,\cite{wang2020minilm, phang2018sentence}, these previous works under-explored the effects of the techniques. 
However, this is the first work to use such techniques with empirical justification for mitigating overfitting problems. 

Among the various ILD objectives, we focus on the two most commonly used distillation objectives: multi-head attention (MHA) and intermediate representations (IR).
% The formula for MHA and IR loss function with layer mapping function $m(\cdot)$ for input as student layer $\ell^{S} \in [1, M]$ and output as teacher layer $m(\ell^{S}) \in [1, L]$ are as follows:
Formally, for the student's layer $\ell^{S} \in [1, M]$, the loss function of MHA and IR are as follows:
\begin{align}
    \mathcal{L}_{\text{MHA}}^{\ell^{S}} &= \frac{1}{A_{h}}\sum_{a=1}^{A_{h}} \text{KLD}(\mathbf{A}^{T}_{m(\ell^{S}), a} \vert\vert \mathbf{A}^{S}_{\ell^{S}, a}) \\
    \mathcal{L}_{\text{IR}}^{\ell^{S}} &= \text{MSE}(\mathbf{H}^{T}_{m(\ell^{S})}, \mathbf{W}^{\text{H}}\mathbf{H}^{S}_{\ell}),
\end{align}
where $m(\cdot)$ is layer mapping function that returns teacher layer $m(\ell^{S}) \in [1, L]$.
Note that KLD and MSE are Kullback-Leibler divergence and mean squared error, respectively. We denote $\mathbf{A}$ and $\mathbf{H}$ as MHA and IR. $T$ and $S$ are superscripts for the teacher and student model, and $a$ and $A_h$ indicate the index and the total number of multi-attention heads, respectively. Note that $\mathbf{W}^{\text{H}}$ is a learnable weight matrix for matching the dimension between representations of the teacher and student.
Consistent with previous studies\,\cite{sun2020mobilebert, jiao2019tinybert}, we observe that sequential training of ILD and original KD\,\cite{hinton2015distilling} shows better than joint training of ILD and original KD.
We conduct an experimental study on sequential training of ILD and original KD from our preliminary experiments.
% From our preliminary experiments, we conduct experimental study on sequential training of ILD and original KD.
All the detailed descriptions of the scope of our empirical study are in Appendix~\ref{app:scope}.
\begin{figure}[t]
    \hspace*{\fill}
    \begin{subfigure}[b]{0.45\linewidth} % 0.4 % 0.23
    \centering
    \includegraphics[width=\linewidth]{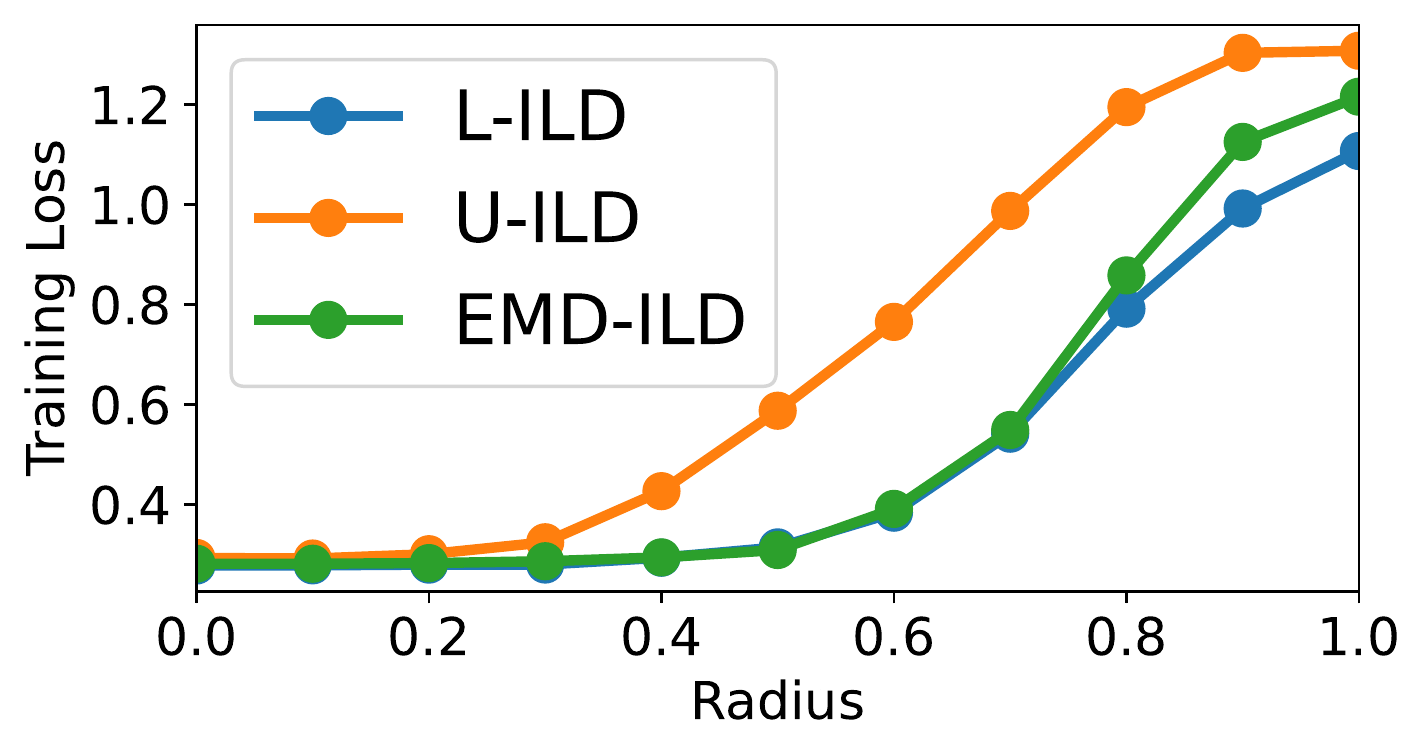}
    \caption{MRPC}
    \end{subfigure}
    \hfill
    \begin{subfigure}[b]{0.45\linewidth} % 0.4 % 0.23
    \centering
    \includegraphics[width=\linewidth]{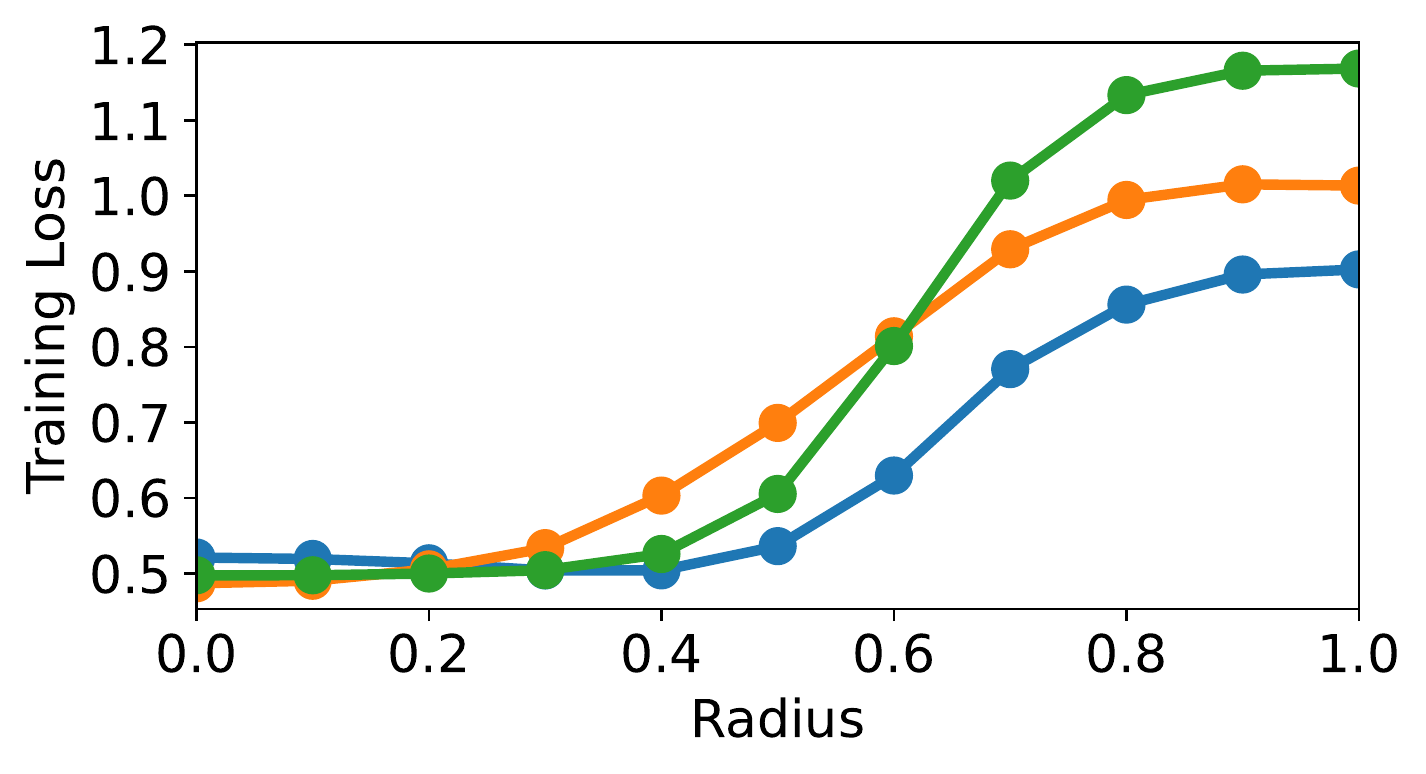}
    \caption{RTE}
    \end{subfigure}
    % \hfill
    % \begin{subfigure}[b]{0.30\textwidth} % 0.4 % 0.23
    % \centering
    % \includegraphics[width=\linewidth]{fig/cr_smooth/QNLI.pdf}
    % \caption{QNLI}
    % \end{subfigure}
    % \hfill
    % \begin{subfigure}[b]{0.24\textwidth} % 0.4 % 0.23
    % \centering
    % \includegraphics[width=\linewidth]{example-image-a}
    % \caption{Class Imbalance}
    % \end{subfigure}
    \hspace*{\fill}
    \caption{Training loss of different distillation approaches\,(L-ILD, U-ILD, EMD-ILD) with increasing Gaussian noise: models trained with L-ILD are more tolerant of noise, which proves that our L-ILD leads models to be more general.}\label{fig:last_robus}
\end{figure}
% While ACs work well on various domains such as self-distillation (Zhang et al. 2019) and class-imbalance (Lee, Shin, and Kim 2021), our work first proposes using ACs with noisy labels based on theoretical motivation. 

\subsection{Layer Mapping: Distill Only the Last Transformer Layer}\label{sec:3.1}
One of the biggest challenges of ILD methods is establishing a proper layer mapping function that determines layers of the teacher and student models to transfer knowledge.
In this section, we observe that transferring layer-to-layer information leads student models to overfit training samples and is the primary reason for the degradation of student performance.
Based on our findings, we suggest that the last layer distillation\,\cite{wang2020minilm, wang2020minilmv2} is promising layer mapping method.
Our empirical analyses can explain the suggested technique's success in terms of mitigating overfitting.
% Our empirical analyses can explain the success of the suggested technique in terms of mitigating overfitting.
% Unlike the previous works\,\cite{wang2020minilm, wang2020minilmv2}, (1) we find that the last layer distillation is much effective in downstream task than pre-training, and (2) our empirical analyses can explain the success of the last Transformer layer distillation while previous works cannot.
\begin{figure*}[t]
    \hspace*{\fill}
    \begin{subfigure}[b]{0.24\textwidth} % 0.4 % 0.23
    \centering
    \includegraphics[width=\linewidth]{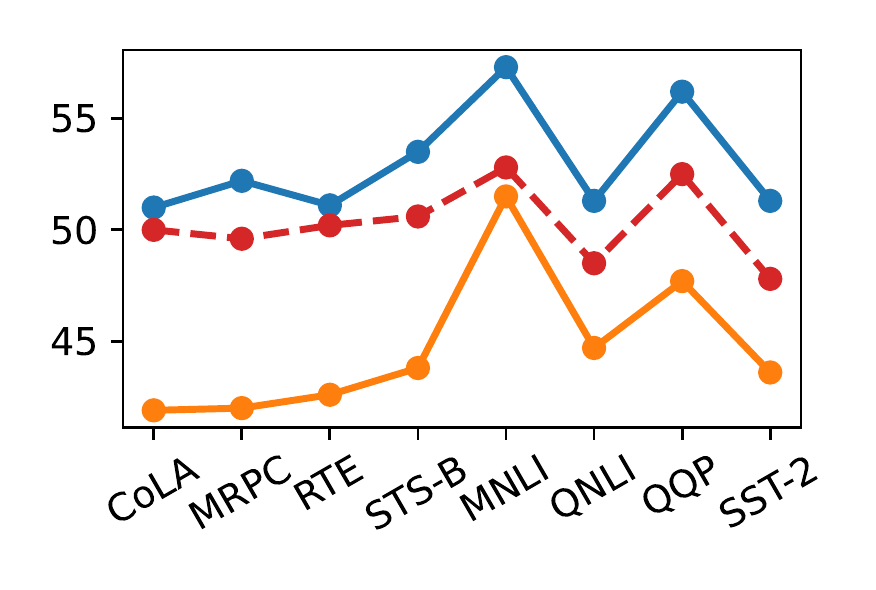}
    \caption{CoLA}
    \end{subfigure}
    \hfill
    \begin{subfigure}[b]{0.24\textwidth} % 0.4 % 0.23
    \centering
    \includegraphics[width=\linewidth]{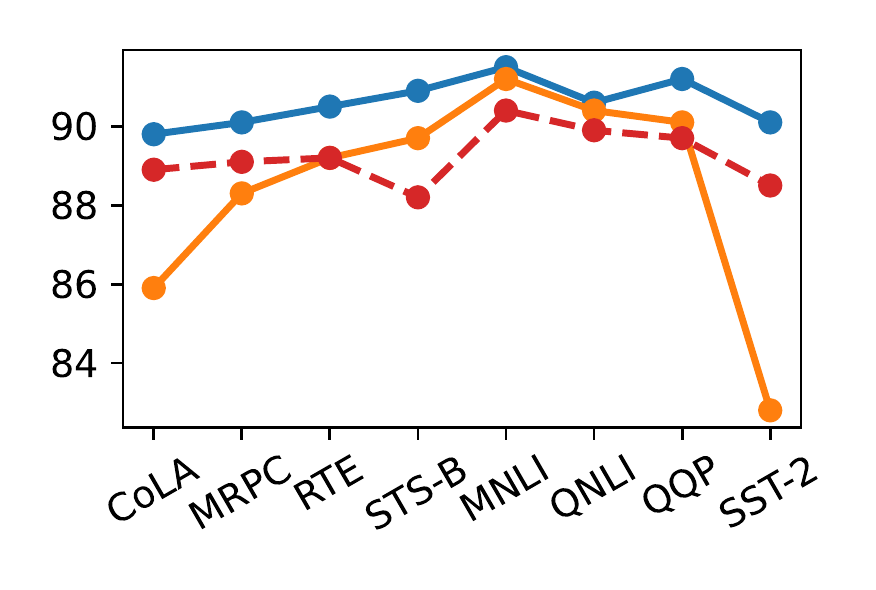}
    \caption{MRPC}
    \end{subfigure}
    \hfill
    \begin{subfigure}[b]{0.24\textwidth} % 0.4 % 0.23
    \centering
    \includegraphics[width=\linewidth]{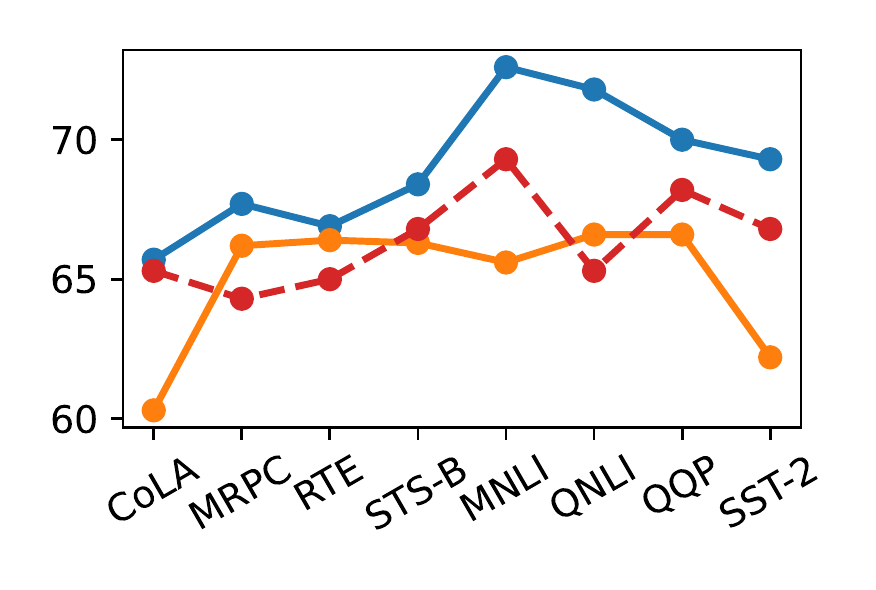}
    \caption{RTE}
    \end{subfigure}
    \hfill
    \begin{subfigure}[b]{0.24\textwidth} % 0.4 % 0.23
    \centering
    \includegraphics[width=\linewidth]{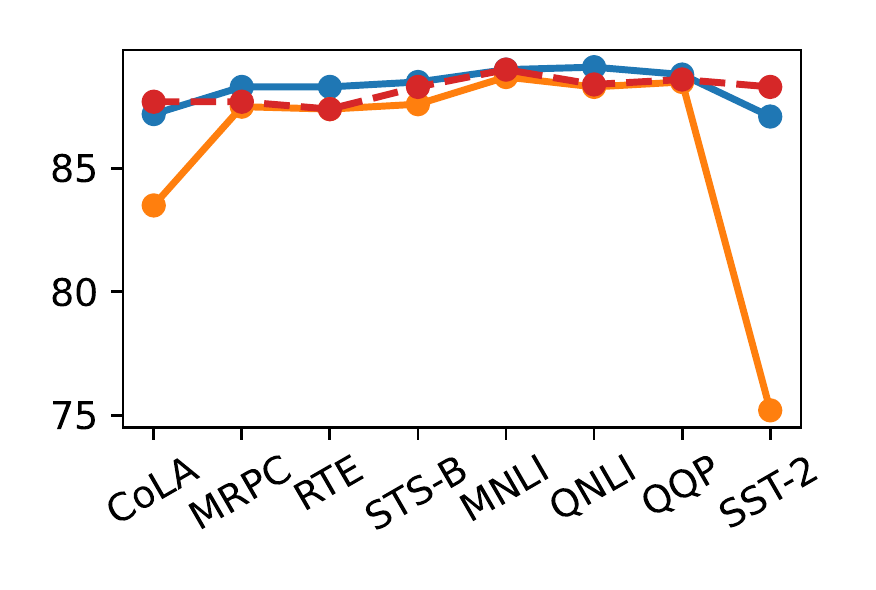}
    \caption{STS-B}
    \end{subfigure}
    \hspace*{\fill}
    \caption{Comparisons for performance of ILD on different supplementary tasks. All students are BERT$_{\text{Small}}$, distilled MHA and IR from BERT$_{\text{BASE}}$ teachers with L-ILD\,(blue) and U-ILD\,(orange). We present the results of prediction layer KD on the supplementary tasks in red dotted lines. All results are averaged over 20 runs.}\label{fig:supp_main}
\end{figure*}
\begin{figure}[t]
    \hspace*{\fill}
    \begin{subfigure}[b]{0.45\linewidth} % 0.4 % 0.23
    \centering
    \includegraphics[width=\linewidth]{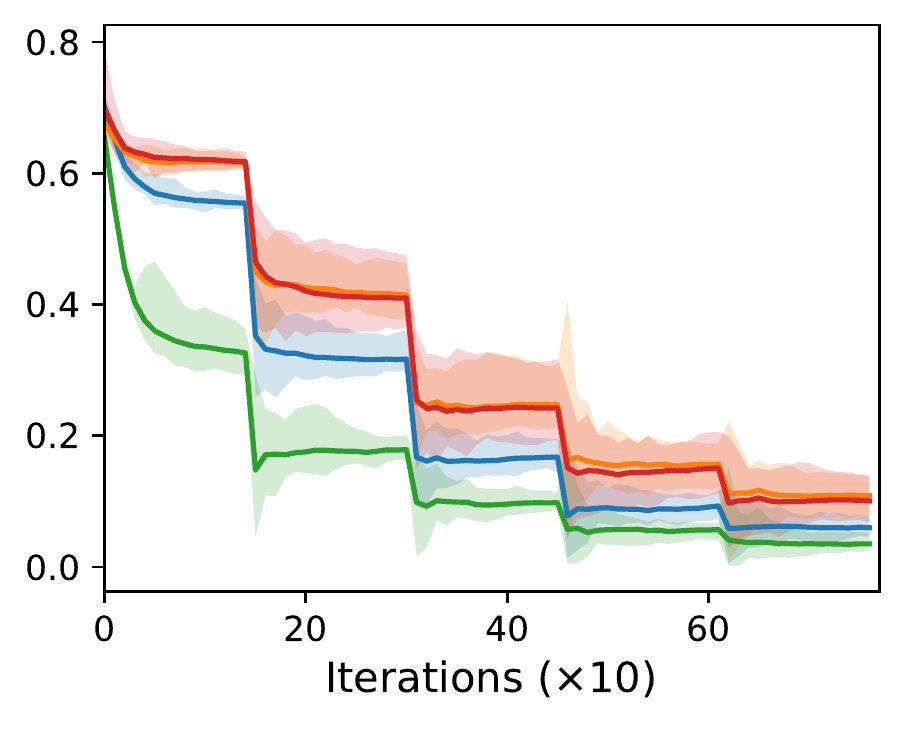}
    \caption{Training}
    \end{subfigure}
    \hfill
    \begin{subfigure}[b]{0.45\linewidth} % 0.4 % 0.23
    \centering
    \includegraphics[width=\linewidth]{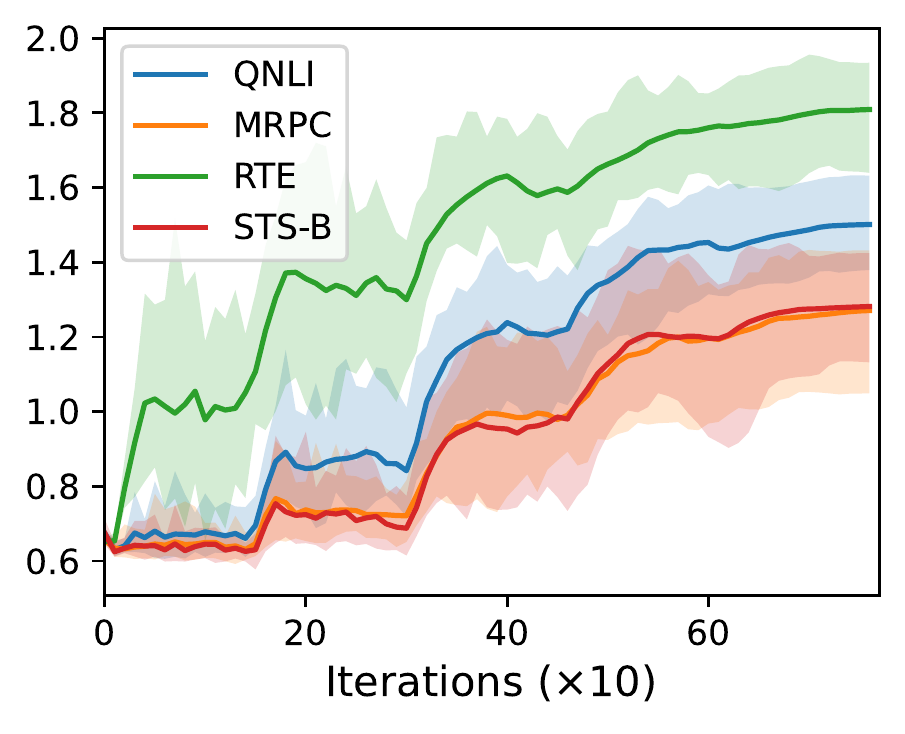}
    \caption{Validation}
    \end{subfigure}
    \hspace*{\fill}
    \caption{The mean\,(solid lines) and range\,(shaded region) of training and validation loss during fine-tuning BERT after conducting ILD on different supplementary tasks, across 20 random trials.}\label{fig:supp_loss}
    \vspace{-10pt}
\end{figure}

\paragraph{Main Observations.}
We compare three distillation strategies: last Transformer layer distillation\,(L-ILD), layer-to-layer distillation using uniform layer mapping\,(U-ILD), and optimal many-to-many layer mapping using the EMD\,(EMD-ILD) proposed in \citet{li2020bert}.
In \autoref{fig:last_main}, L-ILD\,(blue box) outperforms other baselines on all four datasets\,(MRPC, RTE, SST-2, QNLI) in terms of the test performance and variance reduction over the random trials.
Note that the U-ILD, which is a commonly used mapping function\,\cite{sun2019patient, jiao2019tinybert}, leads to performance degradation in most fine-tuning tasks.

We conduct same experiments on the student model with different initialization\,(BERT$_\text{Small}$; \citealt{turc2019well}) as shown in \autoref{fig:last_main2}.
We observe that L-ILD has a higher performance regardless of the size of the dataset or initialized point.
However, the performance gap between L-ILD and other mapping functions gets smaller when the dataset size becomes larger, and the student model is well pre-trained.
On the other hand, although EMD-ILD alleviates the difficulties in layer mapping between the teacher and student, it exhibits lower performance than L-ILD.
We find that performances of EMD-ILD vary across the pre-trained methods while performances of L-ILD are not.
These results validate that the inaccurate layer mapping between the intermediate Transformer layers is not the primary problem of ILD; instead, intermediate Transformer layer distillation itself is the main problem in the fine-tuning stage.

% \paragraph{Smoothness Analysis.}
\paragraph{Analysis.}
To better understand about the performance degradation of distilling the knowledge of intermediate Transformer layers, we evaluate the generalizability of the student models of different layer mapping functions by following \citet{zhang2019your, jeong2020consistency}.
We add Gaussian noise over $\mathcal{N}(0, \sigma^{2}I)$ with different noise radius $\sigma$ to the embedding vectors of the three models\,(L-ILD, U-ILD, EMD-ILD) and then evaluate their cross-entropy loss on the training set.
More generalizable models are robust to the noisy embeddings, hence they have a lower training loss although the magnitude of noise becomes larger.

As shown in \autoref{fig:last_robus}, transferring knowledge of the intermediate Transformer layers leads the student model to the flat minima that are robust of noise and more generalizable\,\cite{hochreiter1997flat, keskar2016large}.
% \paragraph{Loss Surface Analysis.}
% To get further intuition about the performance degradation of distilling the knowledge of intermediate Transformer layers, we provide loss surface visualizations of the U-ILD and L-ILD settings. 
% The parameters of the Truncated BERT, the Last model (student model trained with L-ILD), and the Uniform model (student model trained with U-ILD) are $\theta_0, \theta_1, \theta_2$, respectively. 
% In the subspace spanned by $\delta = \theta_1 - \theta_0$ and $\delta = \theta_2 - \theta_0$, we plot two-dimensional loss surfaces $f(\alpha, \beta) = \mathcal{L}(\theta_0 + \alpha \delta_1 + \beta \delta_2)$ centered on the weights of Truncated BERT $\theta_0$.
% As shown in \autoref{fig:last_surface}, transferring knowledge of the intermediate Transformer layers leads the student model to sharp minima, which results in poorer generalization\,\cite{hochreiter1997flat, keskar2016large}. 
% Thus, the knowledge from the intermediate Transformer layer causes the student model to overfit the training dataset and reduce the generalization.
We further conduct the loss surface\,\cite{zhang2020revisiting} and linear probing\,\cite{aghajanyan2020better} analyses for evaluating the generalizable representations of PLMs during fine-tuning and report the results in Appendix~\ref{app:probe}.
\begin{figure*}[t]
    \hspace*{\fill}
    \begin{subfigure}[b]{0.16\textwidth} % 0.4 % 0.23
    \centering
    \includegraphics[width=\linewidth]{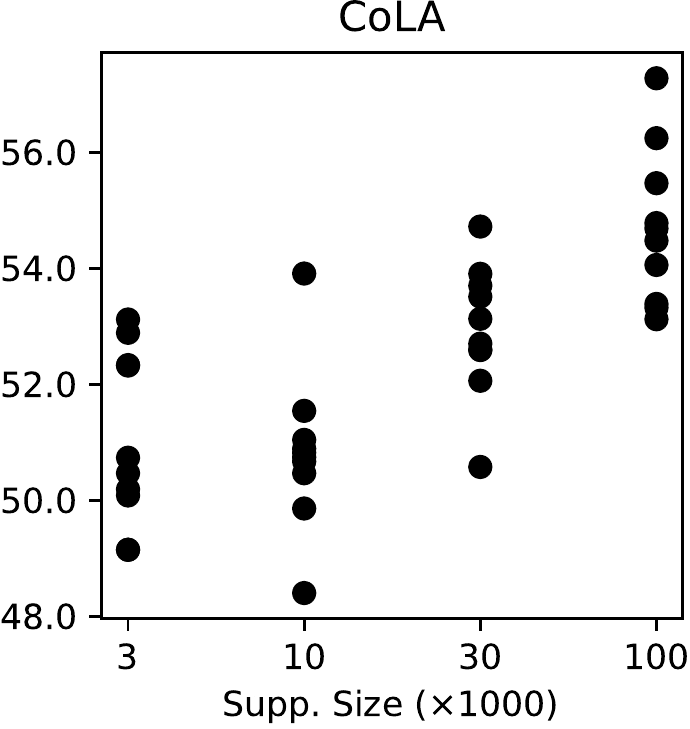}
    \caption{Size\,(CoLA)}\label{fig:supp_task_a}
    \end{subfigure}
    \hfill
    \begin{subfigure}[b]{0.16\textwidth} % 0.4 % 0.23
    \centering
    \includegraphics[width=\linewidth]{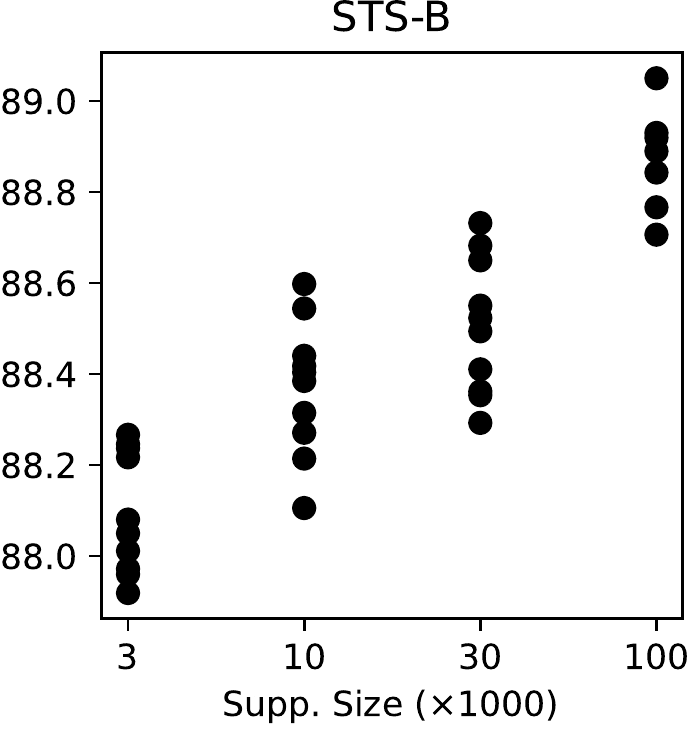}
    \caption{Size\,(STS-B)}\label{fig:supp_task_b}
    \end{subfigure}
    \hfill
    \begin{subfigure}[b]{0.16\textwidth} % 0.4 % 0.23
    \centering
    \includegraphics[width=\linewidth]{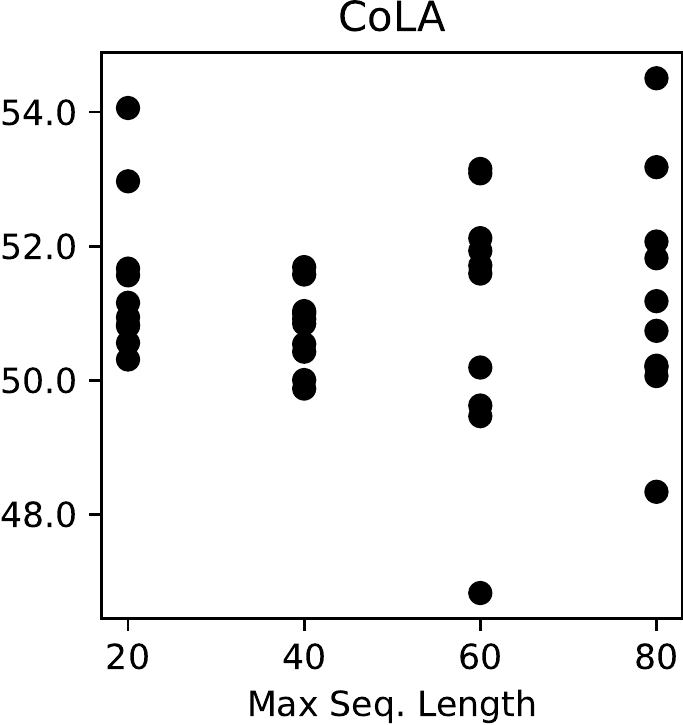}
    \caption{Len.\,(CoLA)}\label{fig:supp_task_c}
    \end{subfigure}
    \hfill
    \begin{subfigure}[b]{0.16\textwidth} % 0.4 % 0.23
    \centering
    \includegraphics[width=\linewidth]{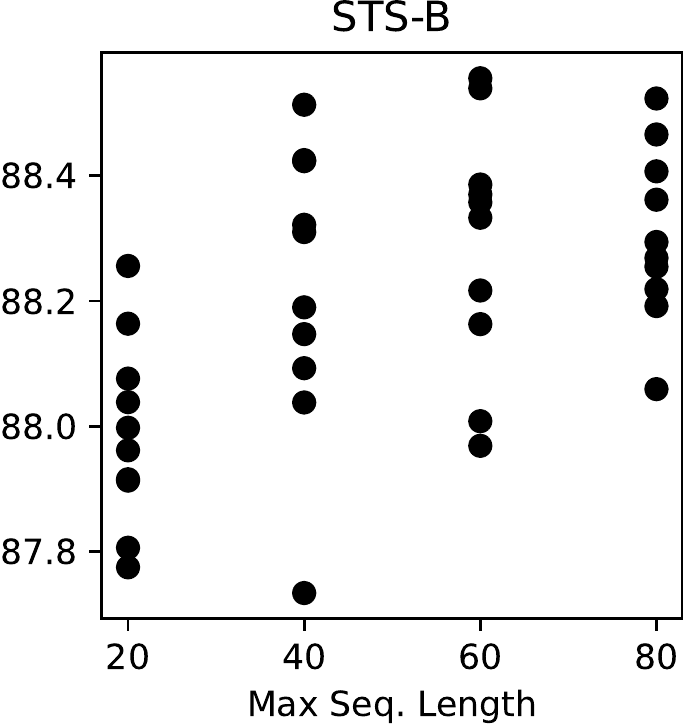}
    \caption{Len.\,(STS-B)}\label{fig:supp_task_d}
    \end{subfigure}
    \hfill
    \begin{subfigure}[b]{0.16\textwidth} % 0.4 % 0.23
    \centering
    \includegraphics[width=\linewidth]{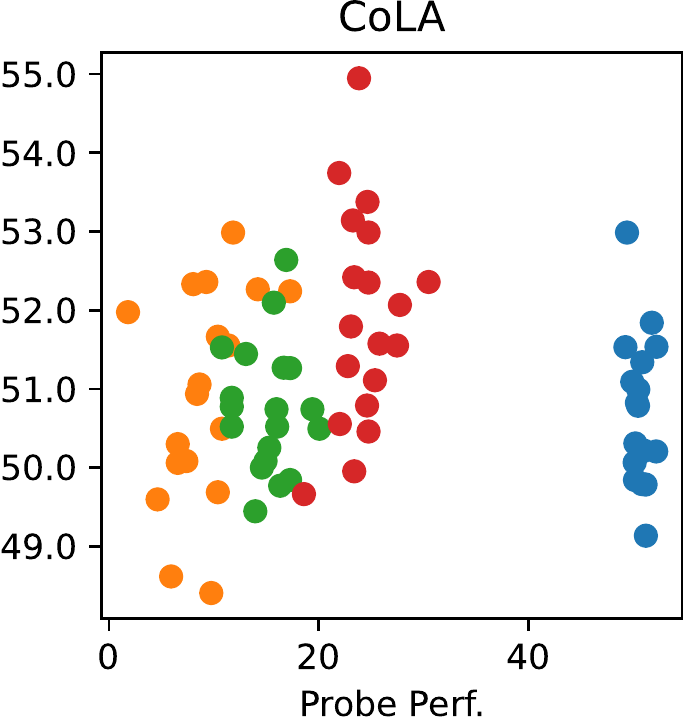}
    \caption{Sim.\,(CoLA)}\label{fig:supp_task_e}
    \end{subfigure}
    \hfill
    \begin{subfigure}[b]{0.16\textwidth} % 0.4 % 0.23
    \centering
    \includegraphics[width=\linewidth]{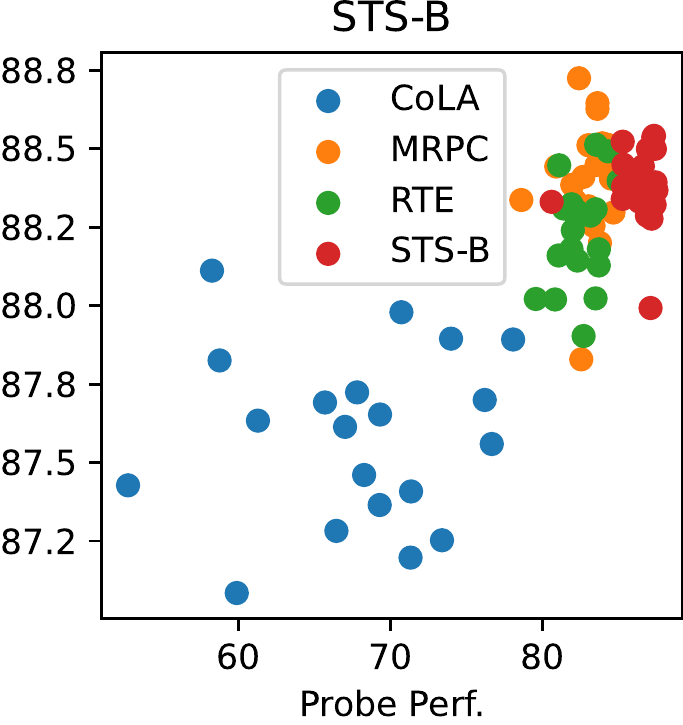}
    \caption{Sim.\,(STS-B)}\label{fig:supp_task_f}
    \end{subfigure}
    \hspace*{\fill}
    \caption{The conditions for appropriate supplementary tasks. The student models trained on supplementary tasks that have large datasets, longer effective sequence length, and high similarity with the target task tend to have higher performances. Size, Len., and Sim. denote the size of the dataset, effective sequence length, and similarity.}\label{fig:supp_task}
    \vspace{-7.5pt}
\end{figure*}

% \paragraph{Linear Probing Analysis.}
% \input{src/last_probing}
% Probing experiments can be used for evaluating the degradation of the generalizable representations of PLMs during fine-tuning. 
% Similar to \citet{aghajanyan2021better}, we conduct the probing method by first freezing the representations from the model trained on one downstream task, and then fine-tuning linear classifiers on top of all Transformer layers to measure the generalization performance of the layers of the teacher and student models.

% Through probing experiments, we observe that the lower-level representations of the Uniform model are overfitted to the training dataset of the target task. 
% XXX shows that the probing performances for 1 to 3 layers of the Uniform model are higher than those of the Last model on the training set of RTE.
% According to \citet{howard2018universal, zhang2020revisiting}, it is crucial to train PLMs so that lower layers have general features and higher layers are specific to target tasks.
% The overfitting of lower layers to the target task leads to performance degradation in the higher layers, as illustrated in XXX.
% Moreover, for the other tasks, the Last models have higher probing performance for all layers than the Uniform models, except for the performance of the first layer on MRPC as indicted in Figures XXX and XXX.
% \input{src/supp_main}
\subsection{Training Data: Use Supplementary Tasks}\label{sec:3.2}
In this section, we investigate the performance of ILD in terms of training datasets for transferring knowledge from teacher to student model.
We observe that conducting ILD even on the last Transformer layer has the risk of overfitting to the training dataset of target task\,(TT).
% From our observation, we find that conducting ILD via supplementary tasks\,(ST) is a simple and efficient method while augmentation module in \citet{jiao2019tinybert} generates 20 times the original data as augmented samples that requires massive computational overhead for both training and generating.
The Previously suggested augmentation module in \citet{jiao2019tinybert} generates 20 times the original data as augmented samples, requiring massive computational overhead for generating.
From our observation, we find that conducting ILD via supplementary tasks\,(ST, \citealt{phang2018sentence}) is a simple and efficient method for overfitting problem.
Based on our observation, we study to find the condition for appropriate ST, which robustly improves the performance of ILD.
% Similar to our work, several studies have been suggested to apply data augmentation for generating 20 times augmented samples from original dataset\,\cite{jiao2019tinybert} XXX XXX XXX 

\paragraph{Main Observations.} 
As shown in \autoref{fig:supp_main}, in most downstream tasks, except for STS-B, the performance of combining ILD with other STs is superior to that when using the original dataset.
Among the tasks with small dataset\,(CoLA, MRPC, RTE, STS-B), although STS-B exhibits superiority as an ST for ILD, all student models with ILD on CoLA exhibit the worst performance for all TTs.
For large tasks\,(MNLI, QNLI, QQP, SST-2) as STs, student models trained on MNLI and SST-2 exhibit the best and worst performance for all TTs.

\paragraph{Analysis.}
To understand performance gain from using STs, we compare the loss dynamics for fine-tuning of RTE task using the cross-entropy loss after conducting ILD on the TT (RTE) and STs (MRPC, STS-B, QNLI).
Notably, the student model with ILD on RTE shows a faster decrease and increase in the training and validation loss, respectively, than the student model with ILD on the STs, as shown in \autoref{fig:supp_loss}.
From the results, we verify that conducting ILD over TT incurs memorization of the student model to training data of TT while performing ILD over ST prevents this memorization yet effectively transfers knowledge of the teacher model.

\subsubsection{Ablation Study}\label{sec:supp_ablation}
Although the combination of ST with ILD generally improves the performances of student models, decreased performances are observed in some cases. These results emphasize the need to select appropriate ST. In this section, we present exploratory experiments on synthetic datasets extracted from the English Wikipedia corpus to provide further intuition for the conditions of convincing STs.

\paragraph{Dataset Size.}
According to the results in \autoref{fig:supp_main}, student models trained on STs with large datasets, such as MNLI and QQP, perform better.
We conducted experiments on synthetic datasets extracted from the Wikipedia corpus with different dataset sizes to validate our observations.
The results in Figure~\ref{fig:supp_task_a} and \ref{fig:supp_task_b} indicate that as the size of the synthetic datasets grows larger, the performance of the student models improves.

\paragraph{Effective Sequence Length.}
A surprising result of \autoref{fig:supp_main} is that ILD on single sentence tasks such as SST-2 or CoLA exhibits lower performances than those of the smaller sentence pair tasks.
This phenomenon is much more evident in U-ILD.
Motivated by these results, we conducted experiments on synthetic datasets with the same dataset size of 30k and different effective sequence lengths (measured without considering [PAD] tokens).
Figures~\ref{fig:supp_task_c} and \ref{fig:supp_task_d} show that as the effective sequence length of the datasets increases, so do the performances of the student models.

\paragraph{Task Similarity.}
Finally, we investigate the effect of task similarity between TTs and STs. 
We only use datasets in the GLUE benchmark for computing the similarity and do not use synthetic Wikipedia datasets. 
To measure the task similarity, we use the probing performance of the TT after performing ILD for each ST, following \citet{pruksachatkun-etal-2020-intermediate}. 
We conduct ILD on different STs and then conduct probing and fine-tuning on the TT. 
Figure~\ref{fig:supp_task_e} and \ref{fig:supp_task_f} summarize the correlation between the probing and fine-tuning performances for CoLA and STS-B as the TT.
The fine-tuning performances get better as the probing performances get better, and it is proven that ILD is better when done on an ST that has a high correlation with the TT.
\begin{figure*}[t]
    \hspace*{\fill}
    \begin{subfigure}[b]{0.30\textwidth} % 0.4 % 0.23
    \centering
    \includegraphics[width=\linewidth]{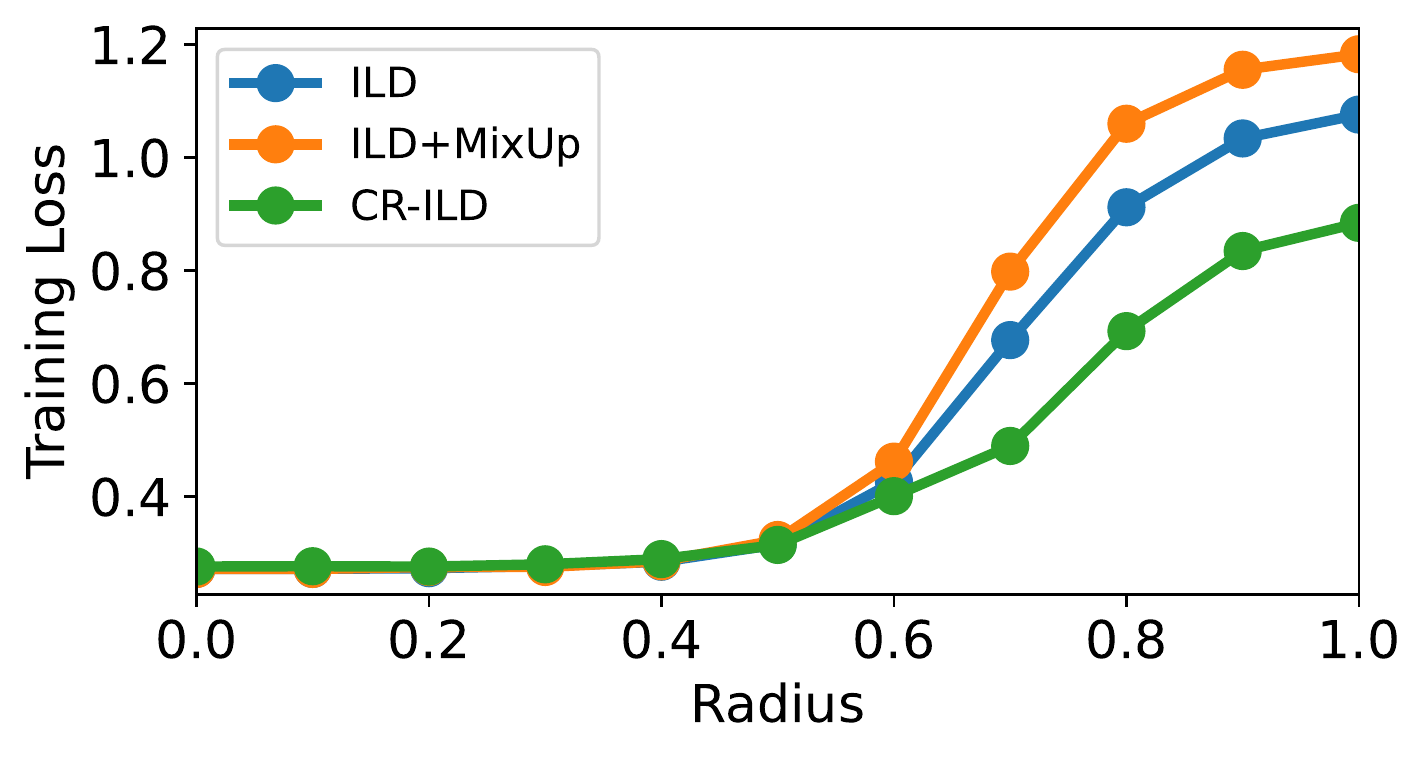}
    \caption{MRPC}
    \end{subfigure}
    \hfill
    \begin{subfigure}[b]{0.30\textwidth} % 0.4 % 0.23
    \centering
    \includegraphics[width=\linewidth]{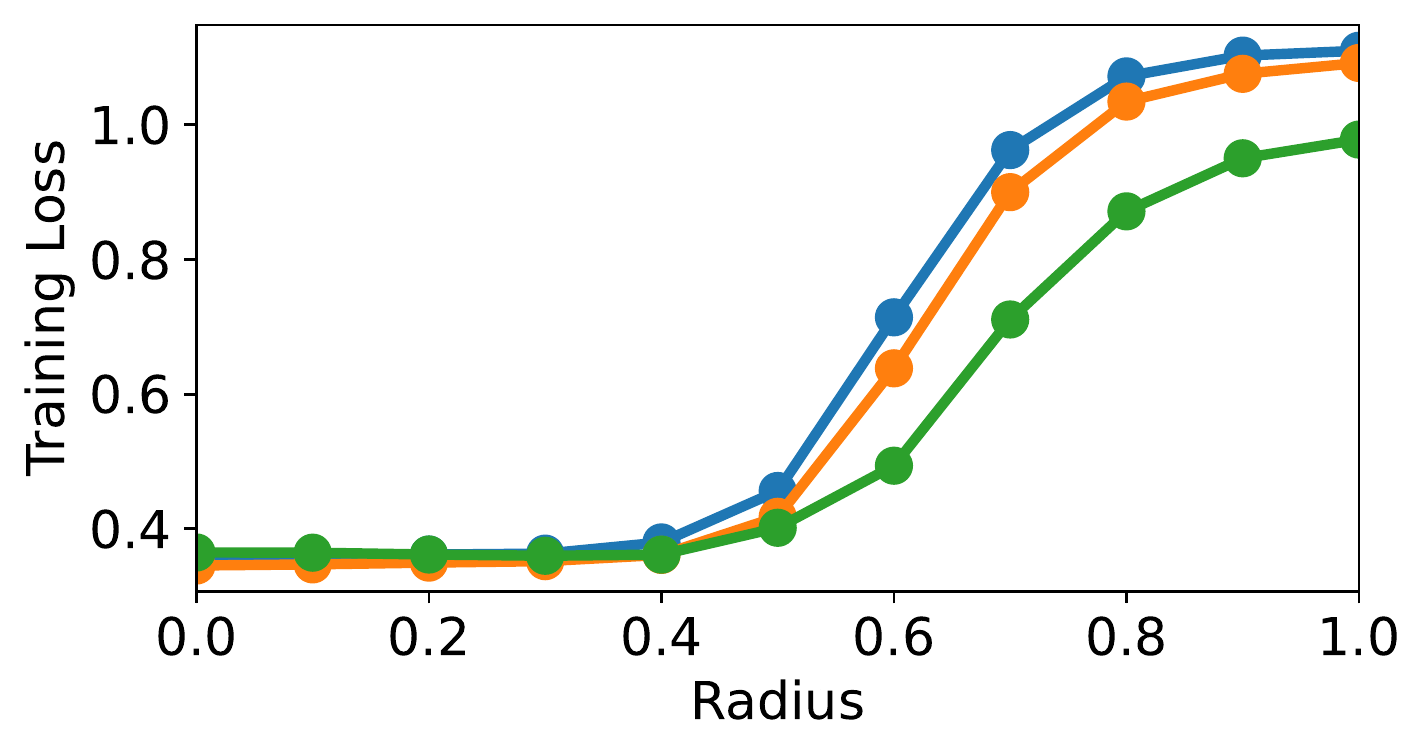}
    \caption{RTE}
    \end{subfigure}
    \hfill
    \begin{subfigure}[b]{0.30\textwidth} % 0.4 % 0.23
    \centering
    \includegraphics[width=\linewidth]{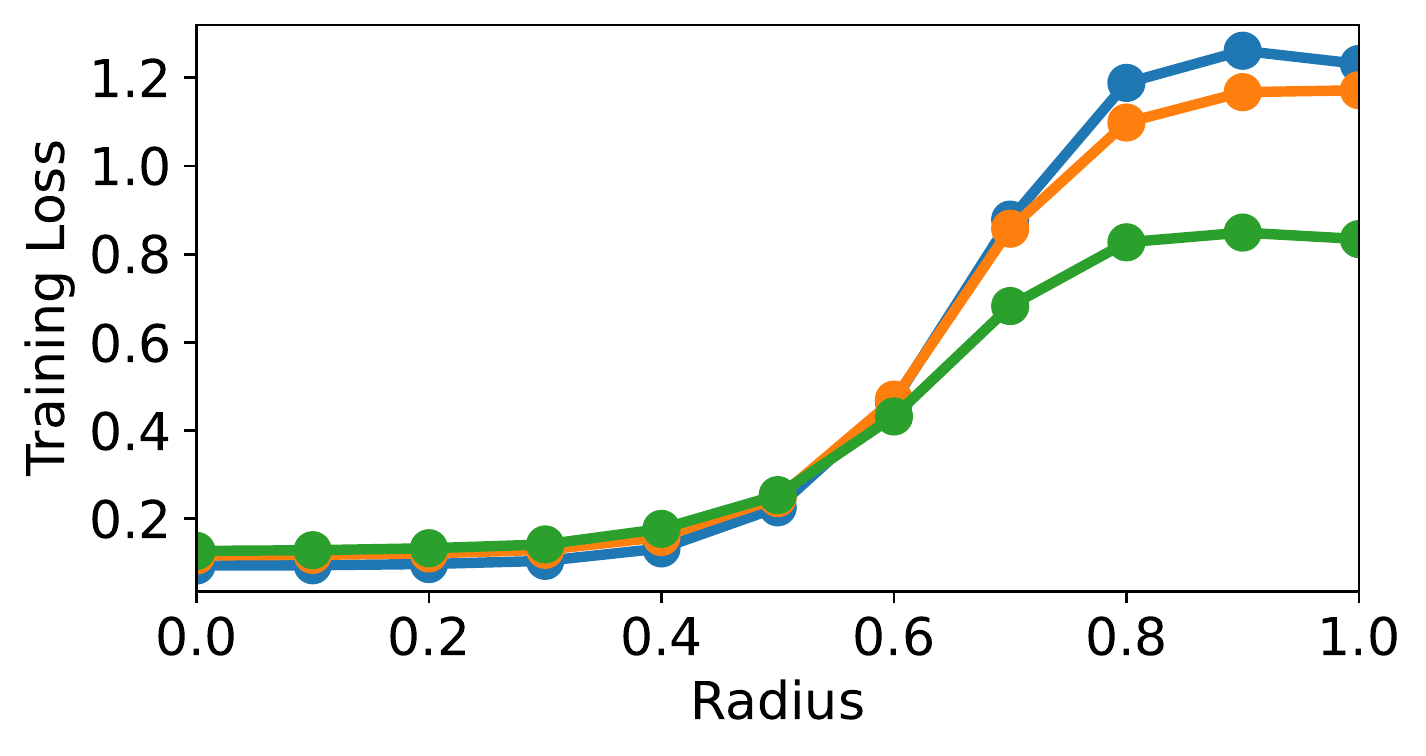}
    \caption{QNLI}
    \end{subfigure}
    % \hfill
    % \begin{subfigure}[b]{0.24\textwidth} % 0.4 % 0.23
    % \centering
    % \includegraphics[width=\linewidth]{example-image-a}
    % \caption{Class Imbalance}
    % \end{subfigure}
    \hspace*{\fill}
    \caption{Comparison of training loss of different distillation approaches (ILD, ILD+MixUp, and CR-ILD) with increasing Gaussian noise: models trained with CR-ILD are more tolerant to noise which verify that our CR-ILD leads model to flat minima which have higher generalization.}\label{fig:cr_smooth}
\end{figure*}

% be your own teacher
% consistency regularization for certified robustness of smoothed classifiers
\section{Method: Consistency Regularized ILD}\label{sec:method}
In this section, we propose a simple yet effective ILD method for improving the robustness of the student models called consistency regularized ILD (CR-ILD) that applies interpolation-based regularization\,\cite{sohn2020fixmatch, zheng-etal-2021-consistency} on MHA and IR of the student models.
Our method efficiently enhances the generalization by leading the student model to the flat minima\,(Section~\ref{sec:3.1}) and introducing appropriate ST\,(Section~\ref{sec:3.2}).
We first introduce the proposed method and then provide analyses of CR-ILD.
\begin{algorithm}[t]
\caption{Consistency Regularized ILD}\label{alg:cr}
\textbf{Input}: embedding layers $\mathbf{W}_{e}^{T}, \mathbf{W}_{e}^{S}$, model parameters $\Theta_{T}, \Theta_{S}$, training dataset $\mathcal{D}$, MixUp hyperparameter $\alpha$, warmup iteration $T$, regularization coefficient $w_{\text{MHA}}^{\text{CR}}, w_{\text{IR}}^{\text{CR}}$ \\
\textbf{Output}: $\Theta_{S}$
\begin{algorithmic}[1]
\State initialize $t \leftarrow 0$
\For{each minibatch B}
\State sample $\vert \text{B} \vert$ pairs of ($\mathbf{x}_{i}, \mathbf{x}_{j}$) for $\mathbf{x}_{i}, \mathbf{x}_{j} \in \text{B}$
\State sample $\lambda \sim \text{Beta}(\alpha, \alpha)$ %, $\lambda'=\max(\lambda, 1-\lambda)$
\State $\mathbf{h}_{i}^{S}=\mathbf{W}_{e}^{S}\mathbf{x}_{i}, \mathbf{h}_{j}^{S}=\mathbf{W}_{e}^{S}\mathbf{x}_{j}$
\State $\mathbf{h}_{i}^{T}=\mathbf{W}_{e}^{T}\mathbf{x}_{i}, \mathbf{h}_{j}^{T}=\mathbf{W}_{e}^{T}\mathbf{x}_{j}$
\State $\tilde{\mathbf{h}}_{i}^{S}=\texttt{Mix}_{\lambda}(\mathbf{h}_{i}^{S}, \mathbf{h}_{j}^{S})$
\State $\tilde{\mathbf{h}}_{i}^{T}=\texttt{Mix}_{\lambda}(\mathbf{h}_{i}^{T}, \mathbf{h}_{j}^{T})$
\State compute $R_{\text{MHA}}$ and $R_{\text{IR}}$ from $\tilde{\mathbf{h}}_{i}^{S}, \mathbf{h}_{i}^{S}, \mathbf{h}_{j}^{S}$
\State compute $\mathcal{L}_{\text{MHA}}$ and $\mathcal{L}_{\text{IR}}$ from $\tilde{\mathbf{h}}_{i}^{S}, \tilde{\mathbf{h}}_{i}^{T}$
\State $\tilde{w}_{\text{MHA}}^{\text{CR}} = \max(\frac{t}{T}, 1) \cdot w_{\text{MHA}}^{\text{CR}}$
\State $\tilde{w}_{\text{IR}}^{\text{CR}} = \max(\frac{t}{T}, 1) \cdot w_{\text{IR}}^{\text{CR}}$
\State $\mathcal{L} \leftarrow \sum_{k \in \{ \text{MHA}, \text{IR}\}} \mathcal{L}_{k}^{M} + \tilde{w}_{k}^{\text{CR}}R_{k}$
\State update $\Theta_{S}$ using gradient descent methods
\State update $t \leftarrow t + 1$
\EndFor
\end{algorithmic}
\end{algorithm}

\subsection{Proposed Method: CR-ILD} % \paragraph{Proposed Method.}
To implement the CR, we apply MixUp\,\cite{zhang2017mixup}, which is an interpolation-based regularizer to improve the robustness in NLP\,\cite{chen2020mixtext}. The direct application of MixUp to NLP is not as straightforward as images, because the input sentences consist of discrete word tokens. Instead, we perform MixUp on the word embeddings at each token by following \citet{chen2020mixtext, liang2020mixkd}. Thus, MixUp samples with embeddings $\mathbf{h}_{i}, \mathbf{h}_{j}$ from sentences $\mathbf{x}_{i}, \mathbf{x}_{j}$ and $\lambda \in [0, 1]$ are generated as:
\begin{align*}
    \texttt{Mix}_{\lambda}(\mathbf{h}_{i}, \mathbf{h}_{j}) = \lambda \cdot \mathbf{h}_{i} + (1-\lambda) \cdot \mathbf{h}_{j},
\end{align*}
Note that $\lambda \sim \text{Beta}(\alpha, \alpha)$ is randomly sampled value from Beta distribution with hyperparameter $\alpha \in (0, \infty)$ for every batch.

Then, we introduce our CR-ILD, as follows:
\begin{align*}
\resizebox{\linewidth}{!}{
    $R_{f_{\theta}} = d(f_{\theta}(\texttt{Mix}_{\lambda}(\mathbf{h}_{i}, \mathbf{h}_{j})), \texttt{Mix}_{\lambda}(f_{\theta}(\mathbf{h}_{i}), f_{\theta}(\mathbf{h}_{j}))),$
}
\end{align*}
where $f_\theta$ denotes the Transformer layer outputs (\textit{e.g.,} MHA and IR) of the model with parameter $\theta$ and embedded input $\mathbf{h}_i, \mathbf{h}_j$. 
Note that $\texttt{Mix}_{\lambda}(f_{\theta}(\mathbf{h}_{i}), f_{\theta}(\mathbf{h}_{j})) = \lambda \cdot f_{\theta}(\mathbf{h}_{i}) + (1-\lambda) \cdot f_{\theta}(\mathbf{h}_{j})$ is interpolation of outputs from $\mathbf{h}_i, \mathbf{h}_j$. 
$d(\cdot, \cdot)$ is a distance metric for regularization, with KLD for MHA and MSE for IR. For example, we have:
\begin{align*}
    R_{\mathrm{MHA}} = \text{KLD}\big(\,&\text{MHA}(\texttt{Mix}_{\lambda}(\mathbf{h}_{i}, \mathbf{h}_{j}))\,||\\
    &\texttt{Mix}_{\lambda}(\text{MHA}(\mathbf{h}_{i}), \text{MHA}(\mathbf{h}_{j}))\,\big) \\[1.5pt]
% \end{align*}
% \begin{align*}
    R_{\mathrm{IR}} = \text{MSE}\big(\,& \text{IR}(\texttt{Mix}_{\lambda}(\mathbf{h}_{i}, \mathbf{h}_{j}))\,,\\
    &\texttt{Mix}_{\lambda}(\text{IR}(\mathbf{h}_{i}), \text{IR}(\mathbf{h}_{j}))\big)
\end{align*}
for CR terms of MHA and IR. 
%In similar, we have CR of IR as follows:
Hence, the overall loss function of CR-ILD is as follows:
\begin{align*}
    \mathcal{L} = \mathcal{L}_{\text{MHA}}^{M} + \mathcal{L}_{\text{IR}}^{M} + \tilde{w}_{\text{MHA}}^{\text{CR}} R_{\text{MHA}} + \tilde{w}_{\text{IR}}^{\text{CR}} R_{\text{IR}},
\end{align*}
where $\tilde{w}_{\text{MHA}}^{\text{CR}}$ and $\tilde{w}_{\text{IR}}^{\text{CR}}$ are coefficients for regularization. As the student models are underfitted to training dataset in the early training phase, we first set the coefficients to zero and gradually increase the values to $w_{\text{MHA}}^{\text{CR}}$ and $w_{\text{IR}}^{\text{CR}}$, respectively. 
Note that both $\mathcal{L}_{\text{MHA}}^{M}$ and $\mathcal{L}_{\text{IR}}^{M}$ are computed by outputs from the teacher and student model with the same MixUp samples as inputs through Eq.\,(1) and Eq.\,(2). All ILD loss and CR term are computed from the last Transformer layer outputs based on Section~\ref{sec:observations}.
% \textcolor{red}{The detailed description of $\mathcal{L}_{\text{MHA}}$ and $\mathcal{L}_{\text{IR}}$ are in Appendix~\ref{app:scope}.} 
We describe the overall algorithm of CR-ILD in Algorithm~\ref{alg:cr}.

\begin{table*}[t]
\centering
\caption{6-layer student results on GLUE development set averaged over 4 runs. $\dagger$ indicates reported results from the \citet{park2021distilling}. Other results are from our re-implementation based on officially released code of original works~\cite{sun2019patient, jiao2019tinybert, li2020bert}.}
\resizebox{0.97\textwidth}{!}{
\begin{tabular}{l|ccc|cccccccc|c}
\toprule
Model & \#Parmas & \#FLOPs & Speedup & CoLA & MNLI & SST-2 & QNLI & MRPC & QQP & RTE & STS-B & AVG \\\midrule
BERT$_\text{BASE}$ & 110M & 22.5B & 1.0x & 59.9 & 84.6 & 92.2 & 91.5 & 90.9 & 91.2 & 70.8 & 89.5 & 83.8 \\ \midrule
\multicolumn{12}{l}{\textit{Truncated BERT\,\cite{sun2019patient} as student model initialization}} \\ \midrule
KD & 67.5M & 11.3B & 2.0x & 36.7 & 82.1 & 90.0 & 88.9 & 89.2 & 90.4 & 65.7 & 88.5 & 78.9 \\
PKD & 67.5M & 11.3B & 2.0x & 37.4 & 82.2 & 90.2 & 89.1 & 89.3 & 90.3 & 66.3 & 87.4 & 79.0 \\
TinyBERT & 67.5M & 11.3B & 2.0x & 31.4 & 81.3 & 89.2 & 86.7 & 87.1 & 90.2 & 57.2 & 84.8 & 76.0 \\
BERT-EMD & 67.5M & 11.3B & 2.0x & 34.6 & 81.5 & 88.5 & 87.9 & 89.1 & 90.2 & 66.4 & 87.9 & 78.3 \\
Ours & 67.5M & 11.3B & 2.0x & \textbf{40.4} & \textbf{82.3} & \textbf{91.1} & \textbf{90.1} & \textbf{89.6} & \textbf{90.7} & \textbf{67.9} & \textbf{89.0} & \textbf{80.1} \\ \midrule
\multicolumn{12}{l}{\textit{BERT$_{\text{Small}}$\,\cite{turc2019well} as student model initialization}} \\ \midrule
KD$^\dagger$ & 67.5M & 11.3B & 2.0x & - & 82.5 & 91.1 & 89.4 & 89.4 & 90.7 & 66.7 & - & - \\
PKD$^\dagger$ & 67.5M & 11.3B & 2.0x & 45.5 & 81.3 & 91.3 & 88.4 & 85.7 & 88.4 & 66.5 & 86.2 & 79.2 \\
TinyBERT$^\dagger$ & 67.5M & 11.3B & 2.0x & 53.8 & 83.1 & 92.3 & 89.9 & 88.8 & 90.5 & 66.9 & 88.3 & 81.7 \\
BERT-EMD & 67.5M & 11.3B & 2.0x & 50.5 & 83.5 & 92.4 & 90.4 & 89.4 & 90.8 & 68.3 & 88.5 & 81.7 \\
CKD$^\dagger$ & 67.5M & 11.3B & 2.0x & 55.1 & 83.6 & \textbf{93.0} & 90.5 & 89.6 & \textbf{91.2} & 67.3 & \textbf{89.0} & 82.4 \\
Ours & 67.5M & 11.3B & 2.0x & \textbf{55.6} & \textbf{83.9} & 92.7 & \textbf{91.4} & \textbf{90.5} & \textbf{91.2} & \textbf{70.2} & 88.8 & \textbf{83.0} \\ \bottomrule
\end{tabular}}\label{tab:glue}
% \vspace{-5pt}
\end{table*}
\subsection{Analysis on CR-ILD} % \paragraph{Analysis.}
In this section, we provide analytical results of CR-ILD to obtain further intuition on our proposed methods. Our CR-ILD regularizes the student model to not learn an undesirable bias by (1) encouraging generalizable student via incurring consistent predictions between MixUp and original samples and (2) generating appropriate ST through MixUp operation.
% \begin{theorem}\label{thm:cr}
%     XXX XXX XXX XXX XXX
% \end{theorem}
% \textcolor{red}{
% \begin{remark}\label{remark:cr}
%     $R_{f_{\theta}}=0$ only holds when $f_{\theta}(\texttt{Mix}_{\lambda}(\mathbf{h}_{i}, \mathbf{h}_{j}))=\texttt{Mix}_{\lambda}(f_{\theta}(\mathbf{h}_{i}), f_{\theta}(\mathbf{h}_{j}))$ for both MHA and IR. If then, the function $f_{\theta}$ is a affine function, i.e., $f_{\theta}(\mathbf{h}) = \mathbf{a}^{\intercal} \mathbf{h} + \mathbf{b}$ for some constant vector $\mathbf{a}, \mathbf{b}$.
% \end{remark}}
% \begin{proof}
%     See Appendix XXX
% \end{proof}
% }
% \textcolor{red}{The Remark~\ref{remark:cr} states that regularization term of CR-ILD leads the function to be affine and smooth for training dataset.}
% \textcolor{blue}{The Remark~\ref{remark:cr} states that the regularization term of CR-ILD leads the function to be affine which is smooth for all data points. This means that CR prevents the overfitting of ILD for the training dataset. We demonstrate the proof for Remark~\ref{remark:cr} and more theoretical analysis on CR-ILD in Appendix~\ref{app:thm}.}

To validate that our CR-ILD makes more generalizable functions empirically, we conduct a similar experiment with \autoref{fig:last_robus} for comparing three models\,(ILD, ILD+MixUp, CR-ILD) as shown in \autoref{fig:cr_smooth}.
ILD+MixUp is the simple combination of ILD and MixUp, which is the same as CR-ILD with $w_{\text{MHA}}^{\text{CR}}$, and $w_{\text{IR}}^{\text{CR}}$ for zero.
% ILD+MixUp is the simple combination of ILD and MixUp, which is the same as CR-ILD with w1, and w2 for zero.
Note that we only use the last Transformer layer for all ILD methods in \autoref{fig:cr_smooth}.
% \textcolor{red}{We add Gaussian noise over $\mathcal{N}(0, \sigma^{2}I)$ with different noise radius $\sigma$ to the embedding vectors of the three models\,(ILD, ILD+MixUp, and CR-ILD) and then evaluate their cross-entropy loss on the training set as shown in \autoref{fig:cr_smooth}.} 
From the results, we obtain that our CR-ILD effectively regularizes the student model not to overfit training data and to be robust to noise injected in embedding spaces. Moreover, it is noteworthy that this smooth regularization is from CR-ILD, whereas the naive application of MixUp does not regularize the student model efficiently. 

Here, we introduce our theoretical analysis that CR-ILD explicitly leads the functions (i.e., MHA, IR) to be convex which is smooth for all data points.
\begin{theorem}[Informal]\label{thm:main}
    Assume that $f_\theta$ satisfies the Assumption~\ref{assum}. With the second order Taylor approximation for $\lambda$ in Definition~\ref{thm:def}, the $\mathcal{L}_{\mathrm{mix}}$ becomes $\hat{\mathcal{L}}_{\mathrm{mix}}$ which can be represented as: 
    \begin{align*}
    &\hat{\mathcal{L}}_{\mathrm{mix}} = \mathcal{L}_{\mathrm{std}} - \frac{2\alpha + 1}{(4\alpha+4)|I|} \sum_{j\in I} \grad \hess^{-1}\grad^\top\,,  \\
    &+ \frac{\alpha+1}{(8\alpha+4)|I|^2} \sum_{i,j\in I } R^{*}(f_{\theta}(\mathbf{h_i}), f_{\theta}(\mathbf{h_j}), \mathbf{y}_{i}, \mathbf{y}_{j})
    \end{align*}
    where $\hess=\mathrm{Hess}_\ell(f_{\theta}(\mathbf{h}_{j}),\mathbf{y}_j)$, and $\grad=D_\ell(f_{\theta}(\mathbf{h}_{j}),\mathbf{y}_j)$. 
    % The detailed form of $R(f_{\theta}(\mathbf{h_i}), f_{\theta}(\mathbf{h_j}), \mathbf{y}_{i}, \mathbf{y}_{j})$ can be found in Appendix~\ref{app:thm}.
\end{theorem}
The detailed form of $R^{*}(f_{\theta}(\mathbf{h_i}), f_{\theta}(\mathbf{h_j}), \mathbf{y}_{i}, \mathbf{y}_{j})$ can be found in Appendix~\ref{app:thm}. Theorem~\ref{thm:main} states that the regularization effect of CR-ILD that makes the significant performance gain of CR-ILD. When we assume that the Hessian can be approximated by the gradient square or outer product of the gradients as in the Gauss-Newton method, the first negative term can be treated as nearly constant. We have the positive term, which performs regularization, and the near-constant negative term. As we discussed earlier, the trainable part of regularizing term reduces the offset related to curvature information. Furthermore, the regularization scheme of CR-ILD can be explained variously. If we assume that the set of data has a non-empty interior, $f(\mathbf{h})$ becomes a linear function, therefore, we can say there is a trend that the function is regularized as a simple smooth function.

% }
% Additionally, we provide the theoretical analysis, that CR-ILD explicitly leads the functions (i.e., MHA, IR) to be convex which is smooth for all data points, in the Appendix.
\begin{table*}[t]
\centering
\caption{The performance averaged over 4 runs on the GLUE development set of 6-layer student models, which were trained on a 1k down-sampled GLUE training set or a GLUE training set under symmetric label noise. We use officially released codes for the re-implementation of PKD\,\cite{sun2019patient}, TinyBERT\,\cite{jiao2019tinybert}, and BERT-EMD\,\cite{li2020bert}. For label noise experiments, we do not consider STS-B for computing average values.}
\resizebox{0.97\textwidth}{!}{
\begin{tabular}{l|ccc|cccccccc|c}
\toprule
Model & \#Parmas & \#FLOPs & Speedup & CoLA & MNLI & SST-2 & QNLI & MRPC & QQP & RTE & STS-B & AVG \\\midrule
% BERT$_\text{BASE}$ & 110M & 22.5B & 1.0x & 59.9 & 84.6 & 92.2 & 91.5 & 90.9 & 91.2 & 70.8 & 89.5 & 83.8 \\ \midrule
\multicolumn{12}{l}{\textit{1k down-sampled\,\cite{zhang2020revisiting} for few-samples experiments}} \\ \midrule
BERT$_{\text{BASE}}$ & 110M & 22.5B & 1.0x & 41.6 & 61.1 & 85.8 & 80.8 & 88.2 & 75.9 & 66.1 & 87.6 & 73.4 \\ \midrule
KD & 67.5M & 11.3B & 2.0x & 17.6 & 58.0 & 83.4 & 78.9 & 86.2 & 74.8 & 59.6 & 83.9 & 67.8 \\
PKD & 67.5M & 11.3B & 2.0x & 17.7 & 57.8 & 83.8 & 75.2 & 86.3 & 73.9 & 59.1 & 83.4 & 67.2 \\
TinyBERT & 67.5M & 11.3B & 2.0x & 9.3 & 55.5 & 80.2 & 71.7 & 85.2 & 72.0 & 57.8 & 82.1 & 64.2 \\
BERT-EMD & 67.5M & 11.3B & 2.0x & 18.8 & 58.0 & 84.2 & 78.5 & 86.3 & 74.3 & 62.1 & 84.8 & 68.4 \\
Ours & 67.5M & 11.3B & 2.0x & \textbf{20.1} & \textbf{59.6} & \textbf{85.0} & \textbf{80.3} & \textbf{87.2} & \textbf{75.7} & \textbf{63.5} & \textbf{85.8} & \textbf{69.7} \\ \midrule
\multicolumn{12}{l}{\textit{Under the presence of uniform (symmetric) label noise\,\cite{jin2021instance, liu-etal-2022-noise} with 30\% noise rate}} \\ \midrule
BERT$_{\text{BASE}}$ & 110M & 22.5B & 1.0x & 39.6 & 81.7 & 90.4 & 86.4 & 82.3 & 86.3 & 57.0 & - & 74.8 \\ \midrule
KD & 67.5M & 11.3B & 2.0x & 37.3 & 80.3 & 88.4 & 85.6 & 81.3 & 86.1 & 59.6 & - & 74.1 \\
PKD & 67.5M & 11.3B & 2.0x & 36.8 & 80.0 & 87.6 & 85.4 & 81.1 & 86.2 & 56.2 & - & 73.3 \\
TinyBERT & 67.5M & 11.3B & 2.0x & 29.7 & 79.9 & 87.2 & 84.6 & 81.2 & 85.7 & 51.6 & - & 71.4 \\
BERT-EMD & 67.5M & 11.3B & 2.0x & 38.5 & 80.6 & 87.8 & 84.9 & 81.2 & 86.0 & 57.0 & - & 73.7 \\
Ours & 67.5M & 11.3B & 2.0x & \textbf{39.6} & \textbf{81.2} & \textbf{89.1} & \textbf{86.0} & \textbf{82.3} & \textbf{86.9} & \textbf{61.8} & - & \textbf{75.3} \\ \bottomrule
\end{tabular}}\label{tab:overfit}
\vspace{-5pt}
\end{table*}

Moreover, thanks to MixUp\,\cite{zhang2017mixup, liang2020mixkd} operation, we can effectively generate the appropriate ST\,(Section~\ref{sec:supp_ablation}) via:
\begin{itemize}
    \item From the MixUp operation, the possible number of MixUp samples can be increased infinitely with the choice of original samples and $\lambda$. This operation increases the dataset size with high task similarity since the MixUp samples are created from the interpolation of the original target task.
    \item If sentence $\mathbf{x}_{i}$ contains more word tokens than sentence $\mathbf{x}_{j}$, then the extra word embeddings are mixed up with embeddings of [PAD] tokens. This operation lengthens the effective sequence length of the dataset in Section~\ref{sec:supp_ablation}, which improves the performance of ILD.
\end{itemize}
From our analysis, we verify that our proposed CR-ILD can effectively transfer the knowledge of teacher models with less overfitting on the training dataset.
\section{Experiments}\label{sec:experiments}
To verify the effectiveness of CR-ILD, we compare the performance of ours with previous distillation methods on the standard GLUE and ill-conditioned GLUE benchmark. The descriptions for experimental setup are in Appendix~\ref{app:data} and \ref{app:exp}.

\subsection{Main Results}
\paragraph{Standard GLUE.}  Following the standard setup\,\cite{sun2019patient}, we use the BERT$_{\text{BASE}}$ as the teacher and 6-layer Truncated BERT\,\cite{sun2019patient} and BERT$_{\text{Small}}$\,\cite{turc2019well} as the student models. \autoref{tab:glue} summarizes that Ours consistently achieve state-of-the-art performances for almost GLUE benchmark, except for SST-2 and STS-B for BERT$_{\text{Small}}$. Despite the simplicity and efficiency of our proposed method, we obtain strong empirical performance.

\paragraph{Ill-conditioned GLUE.} To verify the robustness of our proposed method, we further conduct the experiments on ill-conditioned GLUE, a synthetic dataset with downsampling or injecting label noise to the GLUE benchmark. Since STS-B is a regression task, we cannot inject noise into STS-B. Hence, we do not consider the STS-B task in label noise experiments. The detailed descriptions for ill-conditioned GLUE are in Appendix~\ref{app:data}. \autoref{tab:overfit}  demonstrate that our proposed method alleviates the overfitting and enhances the performance of the student model under few-samples training datasets or the presence of 30\% of label noise. The results for other noise rate are in Appendix~\ref{app:exp}. The experimental results encourage us to use our method on real-world applications which have a high risk of overfitting on the training datasets. Notably, our proposed method achieve higher performance than the teacher model under the presence of label noise.

\begin{figure}[t]
    \centering
    \includegraphics[width=0.92\linewidth]{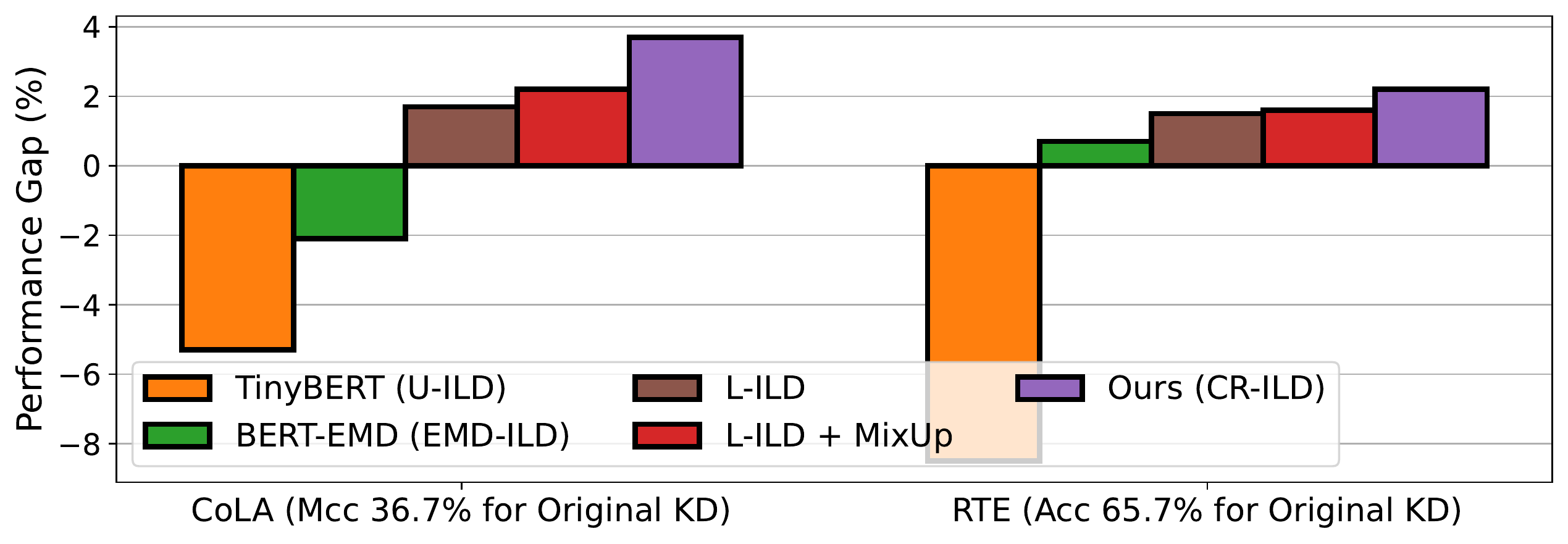}
    \caption{Ablation study on the standard GLUE\,(CoLA, RTE) with Truncated BERT and BERT$_{\text{BASE}}$ as student and teacher models, respectively.}
    \label{fig:exp_ablation}
    \vspace{-7.5pt}
\end{figure}
\subsection{Ablation Study}
To obtain further intuition on CR-ILD, we conduct an ablation study on each component\,(\textit{i.e.,} L-ILD, ST through MixUp, and CR) of our method. Our experiments are conducted on the standard GLUE benchmark with Truncated BERT\,\cite{sun2019patient} as the student models and BERT$_{\text{BASE}}$ as the teacher models. \autoref{fig:exp_ablation} summarizes that all our findings are meaningful, as the performance improves with each addition of a component.
\section{Conclusion}\label{sec:conclusion}
This paper introduces a better use of ILD that transfer knowledge by using outputs of Transformer layers of the teacher and the student models.
We found that existing ILD methods may lead the student model to overfit the training dataset of target tasks and degenerate the generalizability.
Furthermore, we investigated that conducting the ILD (1) only for the last Transformer layer and (2) on supplementary tasks can alleviate the overfitting problems.
Based on our observations, we proposed consistency-regularized ILD that incurs smoother functions and enhance the generalizability of the student models.
Our proposed method effectively distills the knowledge of teacher models by (1) encouraging the flat minima of function from consistency regularization between original embeddings and MixUp embeddings of the student models and (2) efficiently generating appropriate supplementary tasks demonstrated in our findings via MixUp operation.
The experimental results showed that our proposed method could achieve state-of-the-art performance on various datasets, such as the standard and ill-conditioned GLUE benchmarks.

\section*{Limitations}
% 우리 연구는 teacher-student layer mapping의 문제를 다루고 있을 뿐, 모델 내부 layer간, hidden state간 relation이 overfitting에 미치는 영향을 다루고 있지는 않음.
% Our work handles the layer mapping between the teacher and the student networks.
Our work handles the over-fitting of the student network caused by the layer mapping between the teacher and the student networks, which is widely used in \citet{jiao2019tinybert, li2020bert}.
Although we show that our proposed regularization technique can mitigate the over-fitting of the student, the relationship between layers inside the model and the hidden state of tokens in one layer\,\cite{park2021distilling} was not sufficiently considered.
% In addition, We introduce the over-fitting problem of the student network in the knowledge distillation with extensive sentence classification tasks in GLUE Benchmark.
In addition, we back up our proposed idea with theoretical analysis and extensive experiments in sentence classification.
% However, it is somewhat unclear whether our methodology could show the same tendency in other tasks, such as token classification and question answering.
We plan to perform token classification and question-answering experiments to expand our methods to other tasks.
% Based on our observation that the distillation with uniform mapping can cause the over-fitting of the student. 

\section*{Ethics Statement}
Our work complies with all ethical considerations. We hope our work contributes to environmental issues by reducing the computation cost of large PLMs.

\section*{Acknowledgment}
This work was supported by the “Research on model compression algorithm for Large-scale Language Models” project funded by KT (KT award B210001432, 50\%). Also, this work was supported by Institute of Information \& communications Technology Planning \& Evaluation (IITP) grant funded by Korea government (MSIT) [No. 2021-0-00907, Development of Adaptive and Lightweight Edge-Collaborative Analysis Technology for Enabling Proactively Immediate Response and Rapid Learning, 45\%] and [No. 2019-0-00075, Artificial Intelligence Graduate School Program (KAIST), 5\%].

% \bibliography{anthology, custom}
% \clearpage
\bibliography{eacl2023}
\bibliographystyle{acl_natbib}

\clearpage
\appendix
\onecolumn
\begin{center}
    \textbf{\Large{Appendix}} \\
    \textbf{\large{Revisiting Intermediate Layer Distillation for Compressing Language Models: An Overfitting Perspective}}
\end{center}
\section{Theoretical Analysis of CR-ILD}\label{app:thm}
This section gives the theoretical argument that CR-ILD gives additional explicit regularization. We analyze the effect of the MixUp objective function beyond the standard loss function when the CR condition is satisfied. We use the below formulation for objective functions. \textbf{For readability, we partially apply one column style for this section.}
\begin{definition}[Objective Functions]\label{thm:def}
Let us define $\mathcal{D}_\lambda := \mathrm{Beta}(\alpha,\alpha)$, $\tilde{\mathbf{h}}_{ij} := \lambda \mathbf{h}_i + (1-\lambda) \mathbf{h}_j$, and $\tilde{\mathbf{y}}_{ij} := \lambda \mathbf{y}_i + (1-\lambda) \mathbf{y}_j$. Consider the index set $I$. Then the objective functions can be written as:
\begin{align*}
\mathcal{L}_{\mathrm{std}}:= \frac{1}{|I|} \sum_{i\in I }\ell\big(f_{\theta}(\mathbf{h}_{i}),\mathbf{y}_{i}
\big)\quad\mathcal{L}_{\mathrm{mix}}:= \mathbb{E}_{\lambda\sim\mathcal{D}_\lambda}\Big[\frac{1}{|I|^2} \sum_{i,j\in I }\ell\big(f_{\theta}(\tilde{\mathbf{h}}_{ij}),\tilde{\mathbf{y}}_{ij}
\big)  \Big]\,,
\end{align*}
\end{definition}

We assume that CR loss is always optimized during training. That is, if CR loss is $0$, each pair of function values in the loss coincides. Therefore, we can write the first assumption as follows:

\begin{assumption}[Continuation of $f_\theta$ ($\spadesuit$)]
\label{assum}
we assume that $f_\theta$ has continuation on the expanded domain $\{\lambda \mathbf{h}_i + (1-\lambda) \mathbf{h}_j : \lambda\in[0,1]\,,i,j\in I \}$ and the for any convex combination, function value becomes:
\[
f_\theta(\lambda \mathbf{h}_i + (1-\lambda) \mathbf{h}_j) = \lambda f_\theta(\mathbf{h}_i) + (1-\lambda) f_\theta(\mathbf{h}_j)
\]
\end{assumption}
Note also that this continuation can always be well defined if $\{ 
\mathbf{h}_i\}_{i\in I}$ are in general position. Under this assumption, the MixUp loss possesses a regularization effect, which stabilizes the functional outcomes. 

% \newcommand{\vecc}{\mathbf{v}_{\theta,ij}}
%f_{\theta}(\mathbf{h}_{i}) - f_{\theta}(\mathbf{h}_{j})
% \newcommand{\hess}{H_{\ell,j}}
%\mathrm{Hess}_\mathbf{x}\ell(\mathbf{h}_{j},\mathbf{y}_j)
% \newcommand{\grad}{D_{\ell,j}}
%\mathrm{D}_\mathbf{x} \ell(\mathbf{h}_{j},\mathbf{y}_j)

% \onecolumn
\begin{theorem}
\label{CRILDthm}
Assume that $f_\theta$ satisfies the Assumption~\ref{assum}. With the second order Taylor approximation for $\lambda$, the $\mathcal{L}_{\mathrm{mix}}$ becomes $\hat{\mathcal{L}}_{\mathrm{mix}}$ which can be represented as: 
\begin{align*}
\hat{\mathcal{L}}_{\mathrm{mix}} = \mathcal{L}_{\mathrm{std}} &+ \frac{\alpha+1}{(8\alpha+4)|I|^2} \sum_{i,j\in I } \left\Vert (f_{\theta}(\mathbf{h}_{j}),\mathbf{y}_j) - (f_{\theta}(\mathbf{h}_{i}),\mathbf{y}_i) + (2\alpha+1) \hess^{-1}\grad^\top)\right\Vert_{\hess}^2 \\
&- \frac{2\alpha + 1}{(4\alpha+4)|I|} \sum_{j\in I} \grad \hess^{-1}\grad^\top\,,
\end{align*}
where $\hess=\mathrm{Hess}_\ell(f_{\theta}(\mathbf{h}_{j}),\mathbf{y}_j)$, and $\grad=D_\ell(f_{\theta}(\mathbf{h}_{j}),\mathbf{y}_j)$. 
\end{theorem}
Note also that the expectation on higher order of $\lambda$ exponentially decreases as $\mathbb{E}_{\mathcal{D}_\lambda}[\lambda^n] \!\sim\! 2^{-n}$, if $\alpha$ is sufficiently large. The above formulation indicates that the MixUp training with consistency regularization gives further regularization terms, which stabilizes function values $f_\theta(\mathbf{h}_i)$. %That is, we have a regularization term as following:

\subsection{Derivation of the Theorem~\ref{CRILDthm}}
Let us write $\vecc^x=f_{\theta}(\mathbf{h}_{i}) - f_{\theta}(\mathbf{h}_{j})$, $\vecc=(\vecc^x,\mathbf{y}_i - \mathbf{y}_j)$. We first state the second-order Taylor approximation of loss function $\ell$:
\begin{align*}
\mathcal{L}_{\mathrm{mix}}&=\mathbb{E}_{\lambda\sim\mathcal{D}_\lambda}\Big[\frac{1}{|I|^2} \sum_{i,j\in I }\ell\big(f_{\theta}(\tilde{\mathbf{h}}_{ij}),\tilde{\mathbf{y}}_{ij}\big) \Big]\\
&\stackrel{\spadesuit}{=} \mathbb{E}_{\lambda\sim\mathcal{D}_\lambda}\Big[\frac{1}{|I|^2} \sum_{i,j\in I }\ell\big(\lambda f_{\theta}(\mathbf{h}_{i}) + (1-\lambda) f_{\theta}(\mathbf{h}_{j}) ,\lambda \mathbf{y}_i + (1-\lambda) \mathbf{y}_j\big) \Big]\\
&\stackrel{\mathrm{Taylor}}{=} \underbrace{\frac{1}{|I|} \sum_{i\in I } \ell(f_\theta(\mathbf{h}_{i}),\mathbf{y})}_{=:\mathcal{L}_{\mathrm{std}}} + \frac{1}{|I|^2} \sum_{i,j\in I }\Big[ \frac{1}{2}\grad\vecc + \frac{1}{2}\frac{\alpha+1}{4\alpha + 2} \vecc^\top \hess \vecc \Big],
\end{align*}
since $\mathbb{E}_{\lambda\sim\mathcal{D}_\lambda}[\lambda]=1/2$ and $\mathbb{E}_{\lambda\sim\mathcal{D}_\lambda}[\lambda^2] =(\alpha+1)/(4\alpha+2) $. Then,
\begin{align*}
\mathcal{L}_{\mathrm{mix}} & \mathcal{L}_{\mathrm{std}} + \frac{1}{2|I|^2} \sum_{i,j\in I }\Big[  \grad\vecc + \frac{1}{2}\frac{\alpha+1}{2\alpha + 1} \vecc^\top \hess \vecc \Big]\\
&= \mathcal{L}_{\mathrm{std}} + \frac{1}{2|I|^2} \sum_{i,j\in I }\Big[-\frac{2\alpha + 1}{2\alpha + 2} \grad \hess^{-1}\grad^\top +\\
&\qquad\qquad\qquad\qquad\ \frac{1}{2}\frac{\alpha+1}{2\alpha + 1} (\vecc + \frac{2\alpha+1}{\alpha+1} \hess^{-1}\grad^\top )^\top\hess(\vecc + \frac{2\alpha+1}{\alpha+1} \hess^{-1}\grad^\top ) \Big]\\
&= \mathcal{L}_{\mathrm{std}} + \frac{\alpha+1}{(8\alpha+4)|I|^2} \sum_{i,j\in I } \Vert \vecc + (2\alpha+1) \hess^{-1}\grad^\top \Vert_{\hess}^2 - \frac{2\alpha + 1}{(4\alpha+4)|I|} \sum_{j\in I} \grad \hess^{-1}\grad^\top\,.
\end{align*}
\twocolumn
\section{Dataset Description}\label{app:data}

\paragraph{Standard GLUE.} The GLUE benchmark\,\cite{wang2018glue} cover four tasks: natural language inference\,(RTE, QNLI, MNLI), paraphrase detection\,(MRPC, QQP, STS-B), sentiment classification\,(SST-2), and linguistic acceptability\,(CoLA). We mainly focus on four tasks\,(RTE, MRPC, STS-B, CoLA) that have fewer than 10k training samples. While BERT fine-tuning on these datasets is known to be unstable, the ILD on few samples is under-explored. The evaluation metrics for each task of GLUE benchmark are accuracy\,(MNLI, SST-2, QNLI, QQP, RTE), Mcc\,(CoLA), F1 score\,(MRPC), and spearman correlation\,(STS-B). We utilize original split of train, validation\,(development) dataset for our experiments.

\paragraph{Ill-conditioned GLUE.} We use two types of modification on GLUE benchmark, including down-sampling for few-sample GLUE and injecting label noise for corrupted GLUE. For generating few-samples GLUE, we randomly down-sample 1k-sized dataset for each task by following \citet{zhang2020revisiting}. For corrupted GLUE, we follow the experimental setups of \citet{jin2021instance} and inject uniform randomness into a fraction of labels. All other attributes are same for the standard GLUE. Also, we do not modify the development dataset of GLUE benchmark.

\paragraph{Extracted Wiki Corpus in Section~\ref{sec:supp_ablation}} To generate synthetic data, we randomly generate the sample which is consist of two sentences from the Wikipedia corpus (version: enwiki-20200501 from Huggingface). We filter the generated sample by sequence length\,(for experiments of effectiveness of sequence length). We generate new dataset for every single experiment instead of conducting numerous experiment trials to reduce the randomness.
% \section{Scope of Empirical Study in Section~\ref{sec:observations}}\label{app:scope}
% \begin{figure*}[t]
%     \hspace*{\fill}
%     \begin{subfigure}[b]{0.24\textwidth} % 0.4 % 0.23
%     \centering
%     \includegraphics[width=\linewidth]{fig/last_surface/CoLA.png}
%     \caption{CoLA}
%     \end{subfigure}
%     \hfill
%     \begin{subfigure}[b]{0.24\textwidth} % 0.4 % 0.23
%     \centering
%     \includegraphics[width=\linewidth]{fig/last_surface/MRPC.png}
%     \caption{MRPC}
%     \end{subfigure}
%     \hfill
%     \begin{subfigure}[b]{0.24\textwidth} % 0.4 % 0.23
%     \centering
%     \includegraphics[width=\linewidth]{fig/last_surface/RTE.png}
%     \caption{RTE}
%     \end{subfigure}
%     \hspace*{\fill}
%     \caption{last surface.}\label{fig:last_surface}
% \end{figure*}

\begin{figure*}[t]
    \hspace*{\fill}
    \begin{subfigure}[b]{0.24\textwidth} % 0.4 % 0.23
    \centering
    \includegraphics[width=\linewidth]{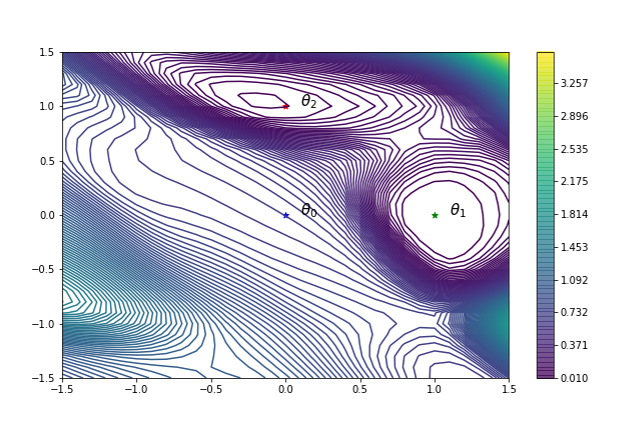}
    \caption{CoLA}
    \end{subfigure}
    \hfill
    \begin{subfigure}[b]{0.24\textwidth} % 0.4 % 0.23
    \centering
    \includegraphics[width=\linewidth]{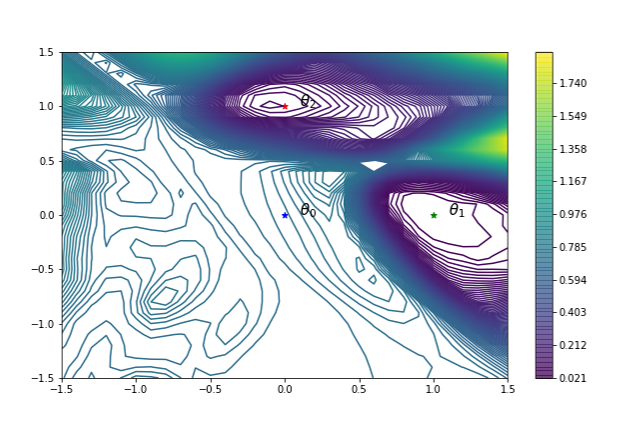}
    \caption{MRPC}
    \end{subfigure}
    \hfill
    \begin{subfigure}[b]{0.24\textwidth} % 0.4 % 0.23
    \centering
    \includegraphics[width=\linewidth]{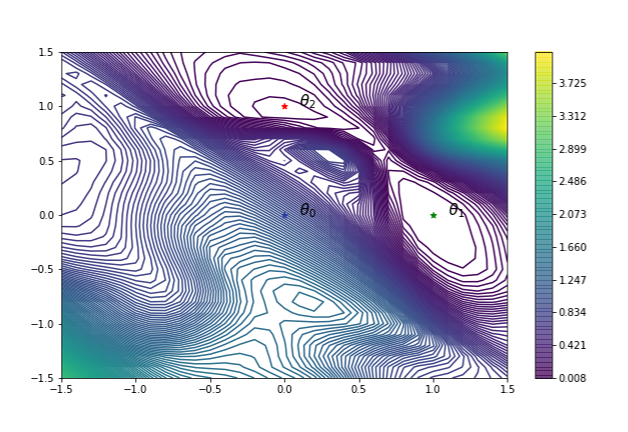}
    \caption{RTE}
    \end{subfigure}
    \hspace*{\fill}
    \caption{2D loss surfaces in the subspace spanned by $\delta_{1} = \theta_{1} - \theta_{0}$ and $\delta_{2} = \theta_{2} - \theta_{0}$ on MRPC and RTE. $\theta_{0}, \theta_{1}, \theta_{2}$ denote the parameters of the Truncated BERT\,(blue), Last model\,(green) and Uniform model\,(red).}\label{fig:last_surface}
\end{figure*}
\begin{figure*}[t]
    \hspace*{\fill}
    % \begin{subfigure}[b]{0.19\textwidth} % 0.4 % 0.23
    % \centering
    % \includegraphics[width=\linewidth]{example-image-a}
    % \caption{RTE\,(teacher, train)}\label{fig:last_probe_a}
    % \end{subfigure}
    % \hfill
    \begin{subfigure}[b]{0.24\textwidth} % 0.4 % 0.23
    \centering
    \includegraphics[width=\linewidth]{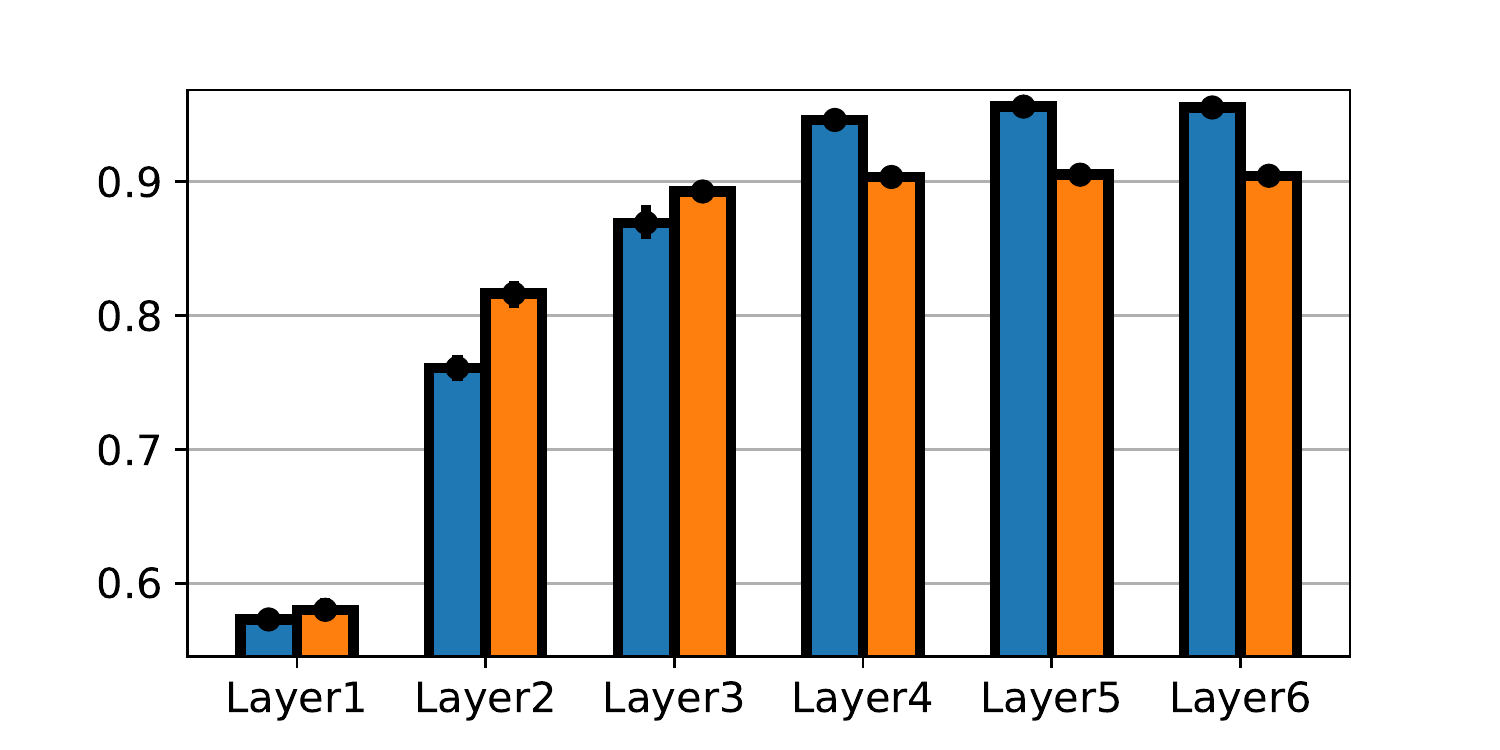}
    \caption{RTE\,(train)}\label{fig:last_probe_b}
    \end{subfigure}
    \hfill
    \begin{subfigure}[b]{0.24\textwidth} % 0.4 % 0.23
    \centering
    \includegraphics[width=\linewidth]{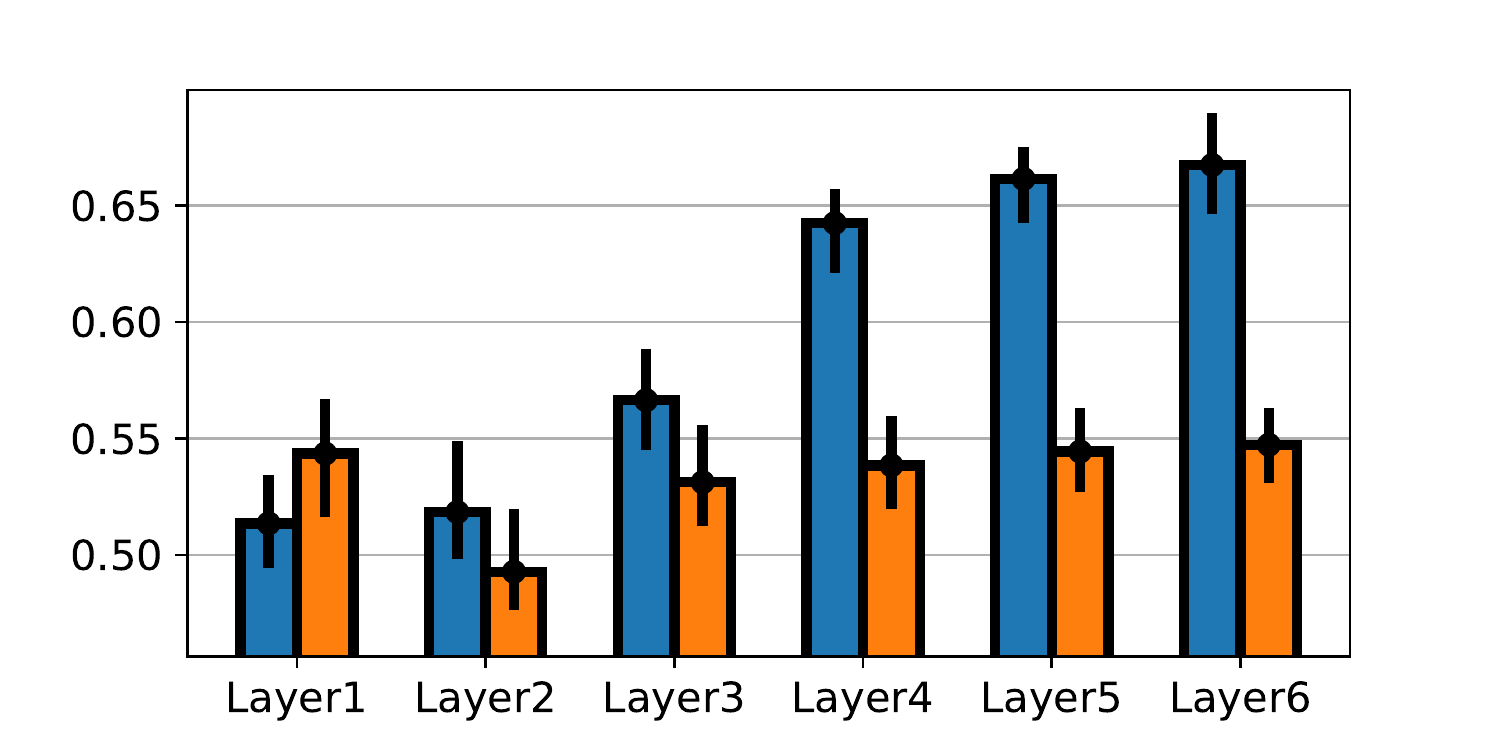}
    \caption{RTE\,(dev)}\label{fig:last_probe_c}
    \end{subfigure}
    \hfill
    \begin{subfigure}[b]{0.24\textwidth} % 0.4 % 0.23
    \centering
    \includegraphics[width=\linewidth]{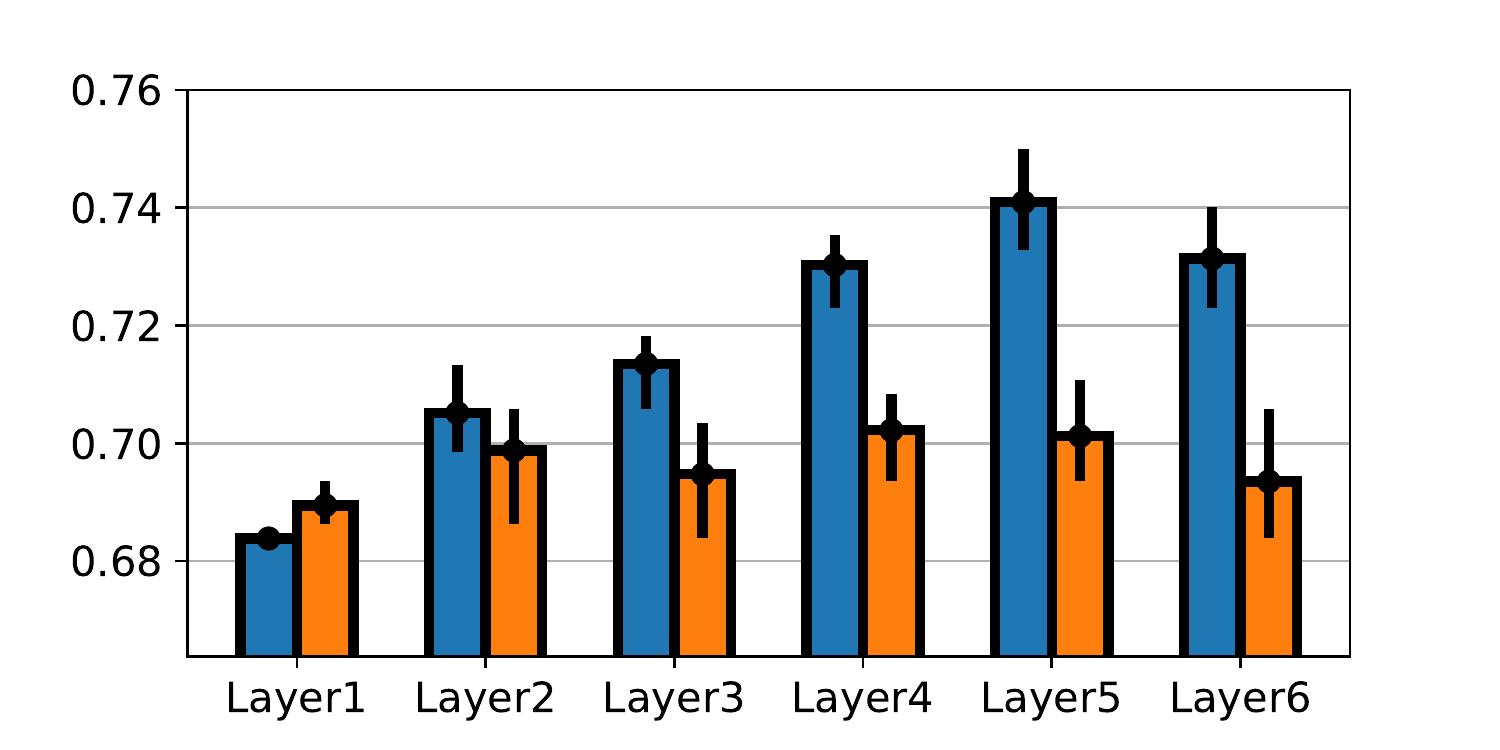}
    \caption{MRPC}\label{fig:last_probe_d}
    \end{subfigure}
    \hfill
    \begin{subfigure}[b]{0.24\textwidth} % 0.4 % 0.23
    \centering
    \includegraphics[width=\linewidth]{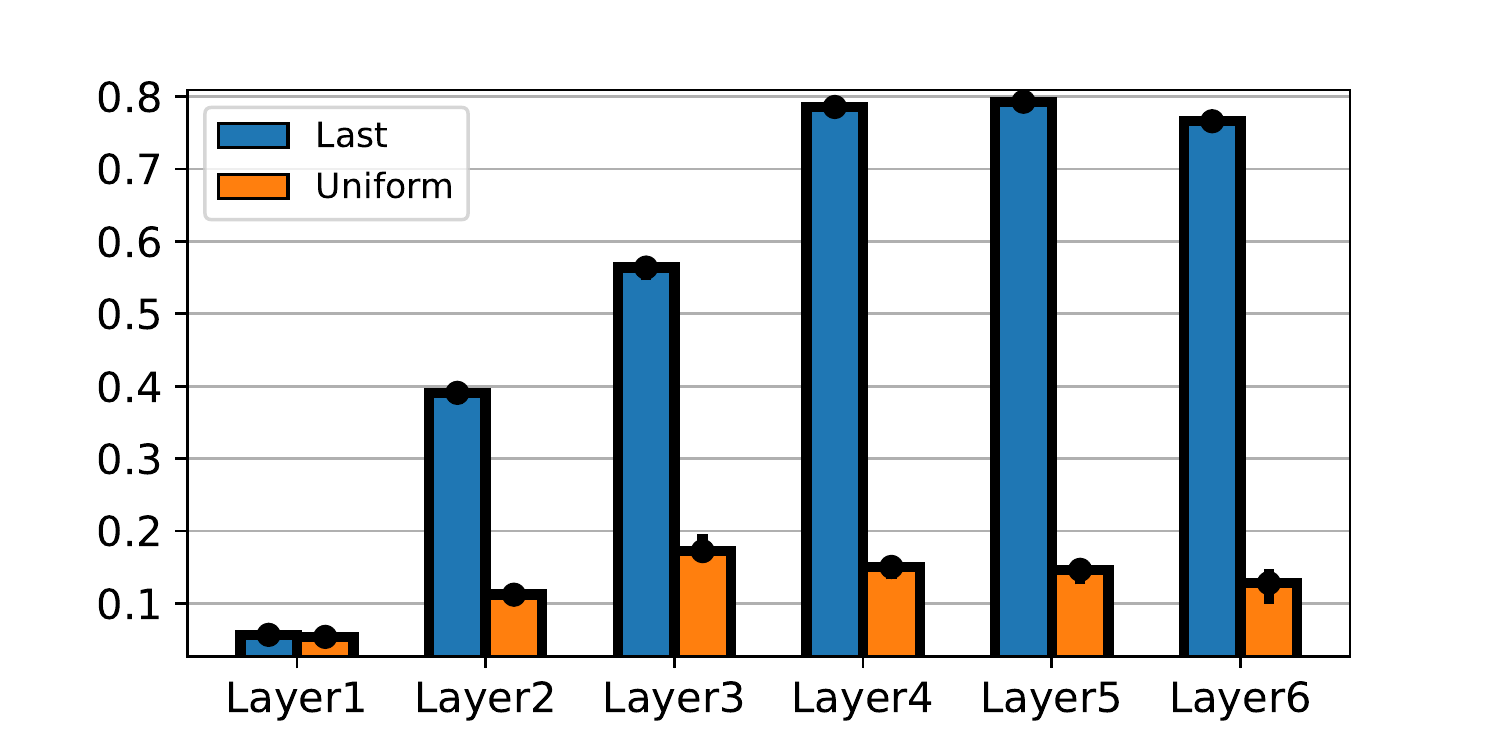}
    \caption{STS-B}\label{fig:last_probe_e}
    \end{subfigure}
    \hspace*{\fill}
    \caption{Results from our layer-wise (x-axis) probe comparing student models trained on RTE with L-ILD and U-ILD., respectively. The student model trained with L-ILD have more generalizable representations than U-ILD.}\label{fig:last_probe}
\end{figure*}
\section{Additional Description for Experiments}\label{app:exp}
\subsection{Scope of Empirical Study in Section~\ref{sec:observations}}\label{app:scope}
\paragraph{Transformer-based Language Models.} Transformer encodes contextual information for input tokens\,\cite{vaswani2017attention}. 
We denote the concatenation of input vectors $\left\{ \mathbf{x}_{i} \right\}_{i=1}^{\vert x \vert}$ as $\mathbf{H}_{0} = \left[ \mathbf{x}_{1}, \ldots, \mathbf{x}_{\vert x \vert} \right]$.
Then, the computation for encoding vectors via stacked Transformer layers is via:
\begin{align*}
    \mathbf{H}_{\ell} = \text{Transformer}_{\ell} (\mathbf{H}_{\ell - 1}), \; \ell \in [1, L].
\end{align*}
The attention mechanism in Transformer improves the performance of NLP significantly and becomes essential. For the $\ell$-th Transformer layer, the output for a self-attention head $ \mathbf{O}_{\ell, a}, a \in \left[ 1, A_h \right]$ is via:
\begin{align*}
    \mathbf{Q}_{\ell, a} = \mathbf{H}_{\ell - 1} \mathbf{W}_{\ell, a}^{Q}, 
    \mathbf{K}_{\ell, a} = \mathbf{H}_{\ell - 1} \mathbf{W}_{\ell, a}^{K}, \\
    \mathbf{A}_{\ell, a} = \text{SoftMax} (\frac{\mathbf{Q}_{\ell, a} \mathbf{K}_{\ell, a}^{\intercal}}{\sqrt{d_k}}), \\
    \mathbf{V}_{\ell, a} = \mathbf{H}_{\ell - 1} \mathbf{W}_{\ell, a}^{V},
    \mathbf{O}_{\ell, a} = \mathbf{A}_{\ell, a} \mathbf{V}_{\ell, a},
\end{align*}
where the previous layer's outputs $\mathbf{H}_{\ell - 1} \in \mathbb{R}^{\vert x \vert \times d_h}$ are linearly projected to a triple of queries, keys, and values using parameter matrices $\mathbf{W}_{\ell, a}^{Q}, \mathbf{W}_{\ell, a}^{K}, \mathbf{W}_{\ell, a}^{V} \in \mathbb{R}^{d_h \times d_k}$, respectively. Note that $A_h$ is the number of attention heads.

\paragraph{Multi-Head Attention.} Many approaches\,\cite{jiao2019tinybert, sun2020mobilebert, wang2020minilm} train the student, making the MHA of the student\,($\mathbf{A}^{S}$) imitate the MHA of the well-optimized teacher\,($\mathbf{A}^{T}$).
\begin{align*}
    \mathcal{L}_{\text{MHA}}^{\ell^{S}} = \frac{1}{A_h} \sum_{a=1}^{A_h} \text{KLD} (\mathbf{A}_{m(\ell^{S}), a}^{T} \vert \vert \mathbf{A}_{\ell^{S}, a}^{S}),
\end{align*}
where KLD is KL-divergence as the loss function. 
Note that $m(\cdot)$ is the layer mapping function for input as student layer $\ell^{S} \in [0, M]$ and output as teacher layer $m(\ell^{S}) \in [1, L]$.
We compare the KLD and mean squared error\,(MSE) for the loss function, and report the results that KLD shows better performance in \autoref{tab:tab_mha}.
\begin{table}[ht]
\centering
\caption{Comparison between KLD and MSE as the loss function for MHA distillation.}\label{tab:tab_mha}
% \footnotesize{
\resizebox{\linewidth}{!}{
\begin{tabular}{l|cccc}
\toprule
   & \textbf{CoLA} & \textbf{MRPC} & \textbf{RTE} & \textbf{STS-B}  \\ \midrule
MHA (KLD) & \textbf{38.1 (1.5)} & \textbf{89.3 (0.5)} & \textbf{67.0 (0.8)} & \textbf{89.1 (0.1)} \\
MHA (MSE) & 37.6 (0.7) & 89.1 (0.5) & 66.5 (1.3) & 89.0 (0.1) \\ \midrule
MHA (KLD) + IR & \textbf{38.4 (1.3)} & \textbf{89.3 (0.3)} & \textbf{67.2 (1.1)} & \textbf{89.1 (0.1)} \\
MHA (MSE) + IR & 38.0 (1.7) & 89.1 (0.3) & 66.3 (0.9) & 89.1 (0.1) \\
\bottomrule
\end{tabular}
}
\end{table}

\paragraph{Intermediate Representation.} Additionally, we study IR, common distillation objective regardless of the network architectures. The MSE between the IR of the teacher\,($\mathbf{H}^{T}$) and student\,($\mathbf{H}^{S}$) is used as the knowledge transfer objective:
\begin{align*}
    \mathcal{L}_{\text{IR}}^{\ell^{S}} = \text{MSE}(\mathbf{H}_{m(\ell^{S})}^{T}, \mathbf{W}^{H} \mathbf{H}_{\ell}^{S}).
\end{align*}
Note that $\mathbf{W}^{H}$ is learnable weight matrix for matching the dimension between representations of the teacher and student. We further compare the IR and patience\,\cite{sun2019patient} in \autoref{tab:tab_mha}.
\begin{table}[ht]
\centering
\caption{Comparisons between Pool and Patience as representation for IR distillation.}\label{tab:tab_ir}
% \footnotesize{
\resizebox{\linewidth}{!}{
\begin{tabular}{l|cccc}
\toprule
   & RTE (RTE) & STS-B (STS-B) & RTE (MNLI) & STS-B (MNLI)  \\ \midrule
Pool  & 60.6 (1.2) & 86.2 (0.3) & \textbf{69.9 (0.8)} & \textbf{89.2 (0.1)} \\
Patience & \textbf{66.2 (1.0)} & \textbf{88.3 (0.4)} & 68.8 (0.5) & 88.8 (0.2) \\ \midrule
Pool + MHA & \textbf{67.2 (1.1)} & \textbf{89.1 (0.1)} & \textbf{70.6 (1.0)} & \textbf{89.6 (0.1)} \\
Patience + MHA & 66.5 (1.5) & 88.4 (0.2) & 68.7 (1.2) & 88.4 (0.2) \\
\bottomrule
\end{tabular}}
\end{table}

\paragraph{Prediction Layer.} The most standard form of KD is logit-based KD\,\cite{hinton2015distilling} for training prediction layer.
\begin{align*}
    \mathcal{L}_{PL} = \text{CE}(\mathbf{z}^{T}/ t, \mathbf{z}^{S}/t).
\end{align*}
We use the cross-entropy\,(\text{CE}) as the loss function with inputs $z^{S}$ and $z^{T}$ as the logit vectors of the student and teacher. We compare the sequential and joint training ILD\,(i.e., MHA, IR) and prediction layer distillation\,(PLD) and report the results that sequential training shows better in \autoref{tab:tab_pld}.
\begin{table}[ht]
\centering
\caption{Comparisons between Sequential and Joint}\label{tab:tab_pld}
% \footnotesize{
\resizebox{\linewidth}{!}{
\begin{tabular}{l|cccc}
\toprule
   & RTE (RTE) & STS-B (STS-B) & RTE (MNLI) & STS-B (MNLI) \\ \midrule
Sequential & \textbf{67.2 (1.1)} &\textbf{ 89.1 (0.1)} & \textbf{70.6 (1.0)} & \textbf{89.6 (0.1)} \\
Joint & 66.7 (1.5) & 88.8 (0.2) & 68.9 (1.4) & 89.3 (0.2) \\
\bottomrule
\end{tabular}}
\end{table}

\section{Experimental Setup}
In this section, we describe the setup for our experimental results. Note that all single experiments are conducted on a single NVIDIA GeForce RTX 2080Ti GPU.

\subsection{Setup for Section~\ref{sec:observations} and Section~\ref{sec:method}}\label{app:set1}
For teacher model, we fine-tune the uncased, 12-layer BERT$_{\text{BASE}}$ model with batch size 32, dropout 0.1, and peak learning rate $2 \times 10^{-5}$ for three epochs. For student model, we mainly use with 6-layer BERT model with initialize point as Truncated BERT \cite{sun2019patient} and BERT$_{\text{Small}}$ \cite{turc2019well}. For fine-tuning student model, under the supervision of a fine-tuned BERT$_{\text{BASE}}$, we firstly perform ILD for 20 epochs with batch size 32 and learning rate $5 \times 10^{-5}$ as follows \citet{jiao2019tinybert}. Then, we conduct prediction layer distillation (PLD) for 4 epochs with choosing batch size 16 and learning rate from $2 \times 10^{-5}$. Unlike the logit-based KD, we only use PLD term and do not use supervision from true labels. while We utilize GLUE\,\cite{wang2018glue} benchmark for exploratory experiments and set the maximum sequence length is set to 128 for all tasks.

\begin{table*}[t]
\centering
\caption{The performance averaged over 4 runs on the GLUE development set of 6-layer student models, which were trained on GLUE training set under symmetric label noise (10\% and 20\%). We use officially released codes for the re-implementation of PKD\,\cite{sun2019patient}, TinyBERT\,\cite{jiao2019tinybert}, and BERT-EMD\,\cite{li2020bert}. For label noise experiments, we do not consider STS-B for computing average values.}
\resizebox{0.93\textwidth}{!}{
\begin{tabular}{l|ccc|cccccccc|c}
\toprule
Model & \#Parmas & \#FLOPs & Speedup & CoLA & MNLI & SST-2 & QNLI & MRPC & QQP & RTE & STS-B & AVG \\\midrule
% BERT$_\text{BASE}$ & 110M & 22.5B & 1.0x & 59.9 & 84.6 & 92.2 & 91.5 & 90.9 & 91.2 & 70.8 & 89.5 & 83.8 \\ \midrule
\multicolumn{12}{l}{\textit{Under the presence of uniform (symmetric) label noise\,\cite{jin2021instance, liu-etal-2022-noise} with 10\% noise rate}} \\ \midrule
BERT$_{\text{BASE}}$ & 110M & 22.5B & 1.0x & 54.0 & 83.1 & 91.1 & 90.0 & 90.6 & 89.7 & 67.5 & - & 80.9 \\ \midrule
KD & 67.5M & 11.3B & 2.0x & 44.9 & 81.6 & 90.6 & 88.9 & 88.7 & 89.6 & 65.0 & - & 78.5 \\
PKD & 67.5M & 11.3B & 2.0x & 45.2 & 81.2 & 90.5 & 89.0 & 89.1 & 89.4 & 65.4 & - & 78.5 \\
TinyBERT & 67.5M & 11.3B & 2.0x & 35.4 & 81.9 & 90.1 & 88.3 & 88.3 & 89.6 & 59.9 & - & 76.2 \\
BERT-EMD & 67.5M & 11.3B & 2.0x & 48.2 & 81.3 & 90.5 & 88.0 & 89.2 & 89.1 & 66.1 & - & 78.9 \\
Ours & 67.5M & 11.3B & 2.0x & \textbf{50.1} & \textbf{82.0} & \textbf{90.7} & \textbf{89.2} & \textbf{89.2} & \textbf{89.6} & \textbf{66.5} & - & \textbf{79.6} \\ \midrule
\multicolumn{12}{l}{\textit{Under the presence of uniform (symmetric) label noise\,\cite{jin2021instance, liu-etal-2022-noise} with 20\% noise rate}} \\ \midrule
BERT$_{\text{BASE}}$ & 110M & 22.5B & 1.0x & 50.8 & 82.4 & 90.0 & 88.6 & 87.7 & 87.9 & 63.2 & - & 78.7 \\ \midrule
KD & 67.5M & 11.3B & 2.0x & 42.7 & 81.5 & \textbf{90.1} & 88.4 & 87.6 & 88.1 & 64.6 & - & 77.6 \\
PKD & 67.5M & 11.3B & 2.0x & 41.8 & 81.4 & 89.4 & 87.9 & 87.5 & 88.0 & 63.0 & - & 77.3 \\
TinyBERT & 67.5M & 11.3B & 2.0x & 31.6 & 81.8 & 89.0 & 87.7 & 87.6 & 88.0 & 56.7 & - & 74.6 \\
BERT-EMD & 67.5M & 11.3B & 2.0x & 40.7 & 81.0 & 89.7 & 87.6 & 88.0 & 87.9 & 64.6 & - & 77.1 \\
Ours & 67.5M & 11.3B & 2.0x & \textbf{44.6} & \textbf{81.9} & 89.8 & \textbf{88.6} & \textbf{88.1} & \textbf{88.2} & \textbf{65.2} & - & \textbf{78.1} \\ \bottomrule
\end{tabular}}\label{tab:other_noise}
\end{table*}
\subsection{Setup for Section~\ref{sec:experiments}}
For achieve higher performance with our methods, we conduct hyper-parameter search as follows:
\begin{itemize}
    \item Peak learning rate\,(ILD): [$2 \times 10^{-5}$, $5 \times 10^{-5}$]
    \item Batch size\,(PLD): [16, 32]
    \item MixUp parameter\,($\alpha$): [0.5, 1.0, 2.0, 3.0]
\end{itemize}
For other hyper-parameter settings are not in the list, we use same parameter values as described in main text or Appendix~\ref{app:set1}. We find that $2 \times 10^{-5}$ is the best peak learning rate of ILD for all tasks except for STS-B. For batch size of PLD stage, RTE, MNLI and QNLI shows higher performance with batch size of 32 and other tasks shows higher performance with batch size of 16. For $\alpha$, a hyperparameter for MixUp operation in CR-ILD, we choose the value of 1.0 by the result of our hyperparameter search. All hyperparameter search are conducted by using \textbf{grid search} with \textbf{averaged three runs}.

\section{Further Experiments on BERT}
\subsection{Further Observation for Section~\ref{sec:3.1}}\label{app:probe}
\paragraph{Loss Surface Analysis.}
To get further intuition about the performance degradation of distilling the knowledge of intermediate Transformer layers, we provide loss surface visualizations of the U-ILD and L-ILD settings. 
The parameters of the Truncated BERT, the Last model (student model trained with L-ILD), and the Uniform model (student model trained with U-ILD) are $\theta_0, \theta_1, \theta_2$, respectively. 
In the subspace spanned by $\delta = \theta_1 - \theta_0$ and $\delta = \theta_2 - \theta_0$, we plot two-dimensional loss surfaces $f(\alpha, \beta) = \mathcal{L}(\theta_0 + \alpha \delta_1 + \beta \delta_2)$ centered on the weights of Truncated BERT $\theta_0$.
As shown in \autoref{fig:last_surface}, transferring knowledge of the intermediate Transformer layers leads the student model to sharp minima, which results in poorer generalization\,\cite{hochreiter1997flat, keskar2016large}. 
Thus, the knowledge from the intermediate Transformer layer causes the student model to overfit the training dataset and reduce the generalization.

\paragraph{Linear Probing Analysis.}
Probing experiments can be used for evaluating the degradation of the generalizable representations of PLMs during fine-tuning. 
Similar to \citet{aghajanyan2021better}, we conduct the probing method by first freezing the representations from the model trained on one downstream task, and then fine-tuning linear classifiers on top of all Transformer layers to measure the generalization performance of the layers of the teacher and student models.

Through probing experiments, we observe that the lower-level representations of the student model related to U-ILD are overfitted to the training dataset of the target task. 
Figure~\ref{fig:last_probe_b} shows that the probing performances for 1 to 3 layers of the student model with U-ILD are higher than those of the Last model on the training set of RTE.
According to \citet{howard2018universal, zhang2020revisiting}, it is crucial to train PLMs so that lower layers have general features and higher layers are specific to target tasks.
The overfitting of lower layers to the target task leads to performance degradation in the higher layers, as illustrated in Figure~\ref{fig:last_probe_c}.
Moreover, for the other tasks, the student models with L-ILD have higher probing performance for all layers than the Uniform models, except for the performance of the first layer on MRPC as indicted in Figure~\ref{fig:last_probe_d} and \ref{fig:last_probe_e}.

\subsection{Experimental Results for Different Label Noise Ratio}
We conduct additional experiments on the GLUE benchmark with different label noise ratios\,(10\% and 20\% of uniform label noise) as shown in \autoref{tab:other_noise}. While BERT-EMD\,\cite{li2020bert} shows the second best performance in small noise ratio\,(10\%) and achieve better performance than the original KD, the original KD and PKD\,\cite{sun2019patient} present the higher performance in severe noise rate (20\% in \autoref{tab:other_noise} and 30\% in \autoref{tab:overfit}) than BERT-EMD. Surprisingly, our CR-ILD\,(Ours) shows the best performance for all noise ratios consistently which verifies that our proposed method encourages the distilling of the knowledge effectively and prevents overfitting on the training datasets.
\section{Further Experiments on Encoder-Decoder Models}

\begin{figure*}[t]
    \hspace*{\fill}
    \begin{subfigure}[b]{0.19\textwidth} % 0.4 % 0.23
    \centering
    \includegraphics[width=\linewidth]{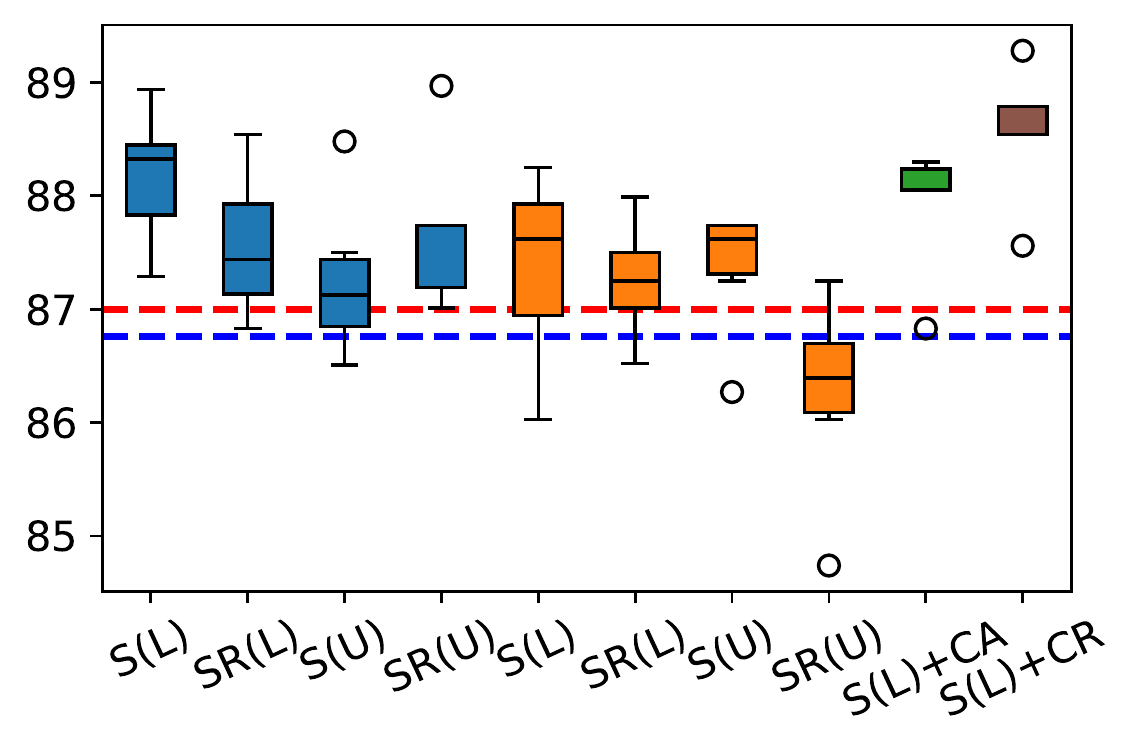}
    \caption{MRPC} \label{fig:appx_t5_a}
    \end{subfigure}
    \hfill
    \begin{subfigure}[b]{0.19\textwidth} % 0.4 % 0.23
    \centering
    \includegraphics[width=\linewidth]{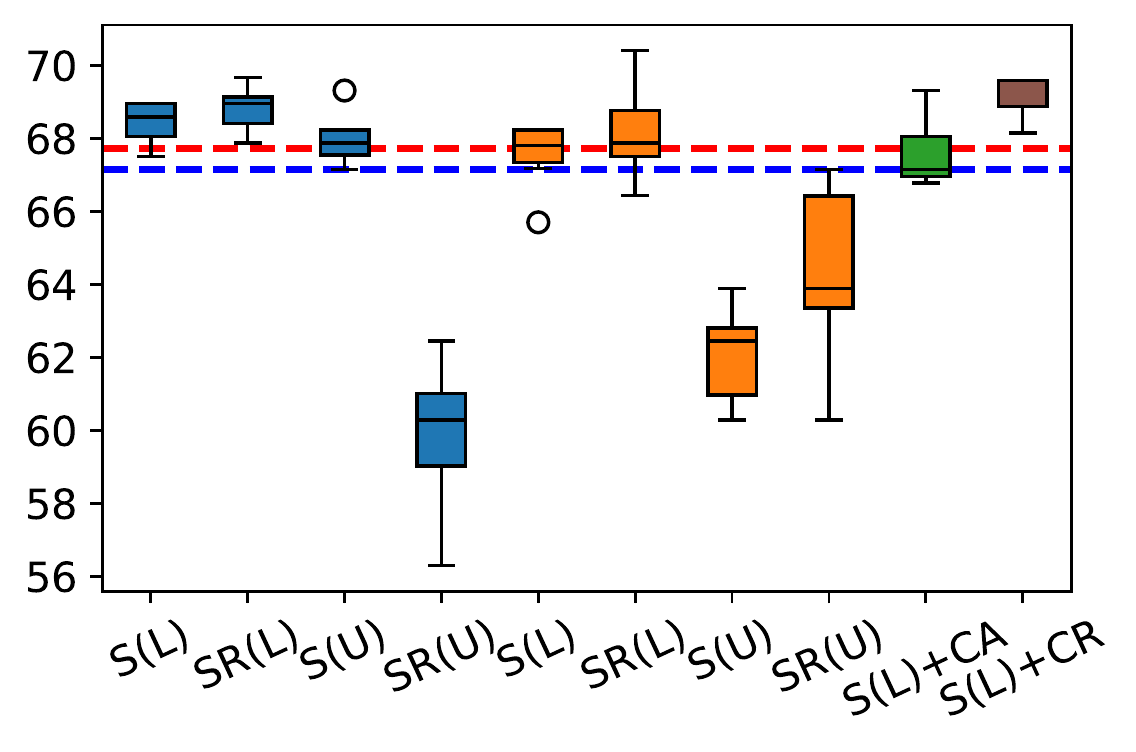}
    \caption{RTE} \label{fig:appx_t5_b}
    \end{subfigure}
    \hfill
    \begin{subfigure}[b]{0.19\textwidth} % 0.4 % 0.23
    \centering
    \includegraphics[width=\linewidth]{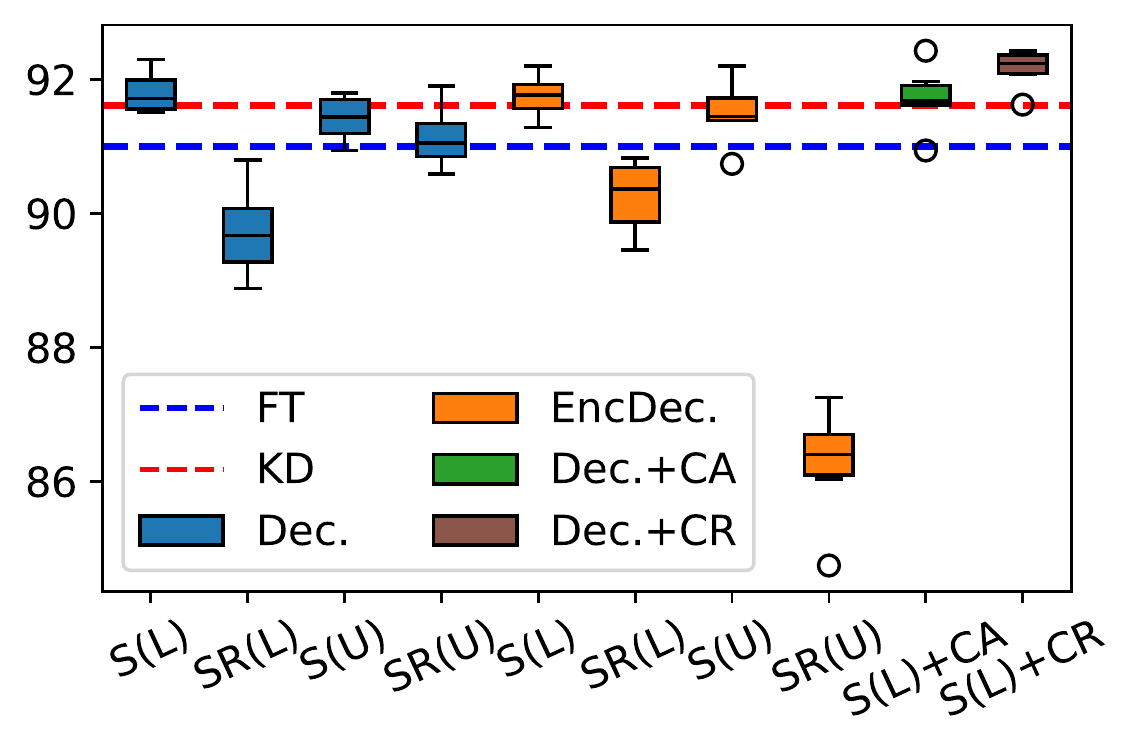}
    \caption{SST-2} \label{fig:appx_t5_c}
    \end{subfigure} 
    \hfill
    \begin{subfigure}[b]{0.19\textwidth} % 0.4 % 0.23
    \centering
    \includegraphics[width=\linewidth]{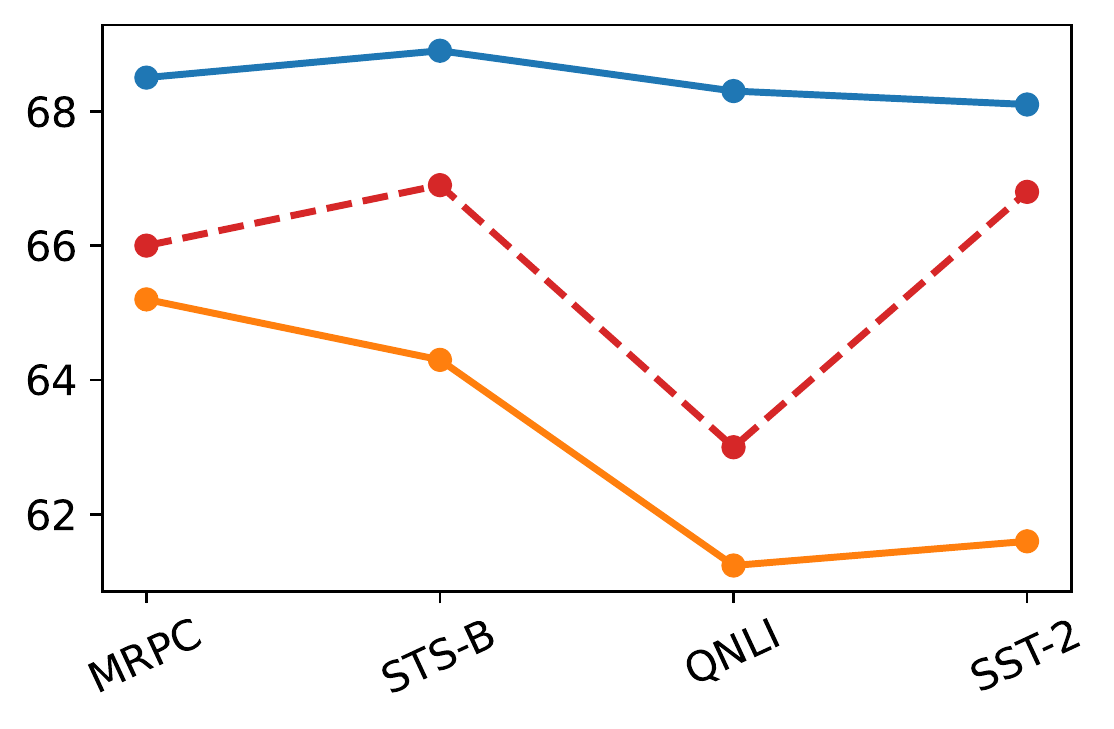}
    \caption{Supp RTE} \label{fig:appx_t5_d}
    \end{subfigure}
    \hfill
    \begin{subfigure}[b]{0.19\textwidth} % 0.4 % 0.23
    \centering
    \includegraphics[width=\linewidth]{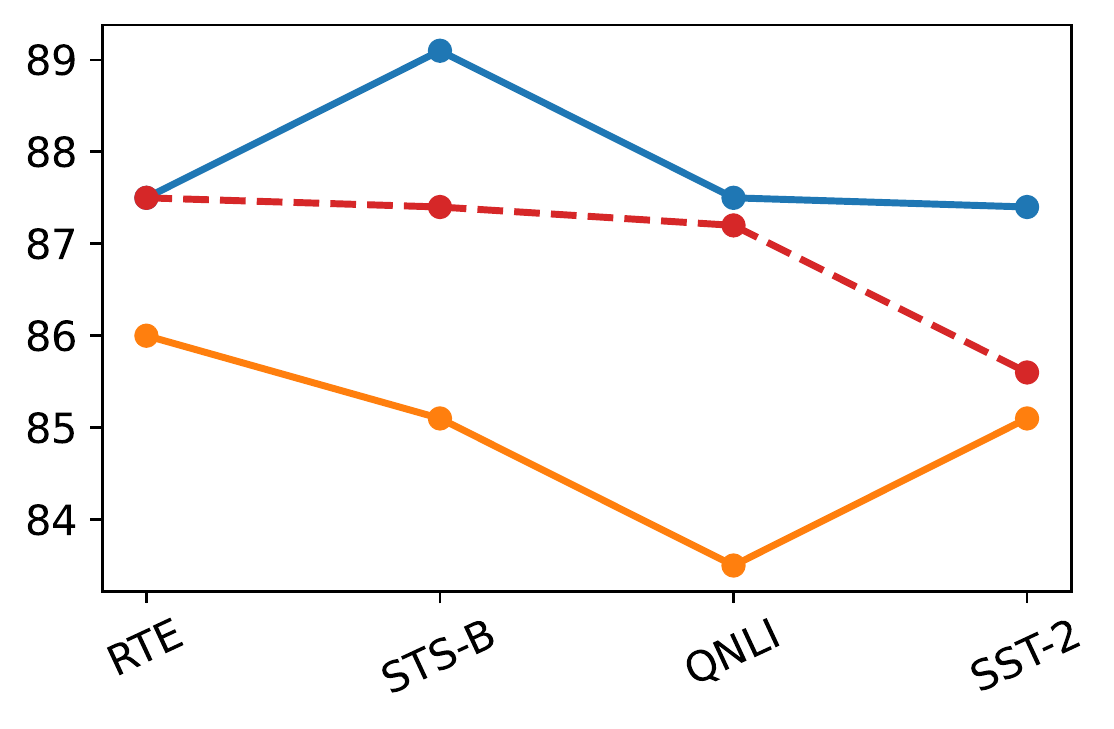}
    \caption{Supp MRPC} \label{fig:appx_t5_e}
    \end{subfigure}
    % \hfill
    % \begin{subfigure}[b]{0.19\textwidth} % 0.4 % 0.23
    % \centering
    % \includegraphics[width=\linewidth]{fig/appx_t5/T5_2d_loss.pdf}
    % \caption{Sim.\,(STS-B)}
    % \end{subfigure}
    % \hspace*{\fill}
    \caption{We compare the performance for different layer mapping functions and distillation objectives in (a)-(c). The distillations of the last layer, uniform layer mapping, self-attention, and IR are denoted by L, U, S, and R, respectively. The blue\,(Enc.) and orange\,(Dec.) bar denote the application of ILD to the decoder network only and to both encoder and decoder network, respectively. We also denote ILD with CA and CR as the green and brown bars in (a)-(c), respectively. (d)-(e) are results for using ST for TTs of MRPC and RTE. We only distill self-attention of the last layer of the decoder when using ILD with ST and CR}\label{fig:appx_t5}
\end{figure*}

% caption 다시 적을 것!
\subsection{T5: Study on Encoder-Decoder Models}
% Some previous works handled the compression of Transformer model including decoder architecture. 
% \cite{li2021short} applied previous KD methods \cite{jafari2021annealing, rashid2021mate, haidar-etal-2022-rail} already shown as an effective way for the Transformer encoder-based model to the decoder-based model, DistilGPT-2 \textcolor{red}{cite!}.
% \cite{}
In this section, we apply our approaches to T5 to generalize our result from the encoder-based model to the encoder-decoder model.
First, we explain our experimental setup in the experiments conducted on T5.
Secondly, we examine (1) two findings (last Transformer layer, supplementary task) and (2) our proposed method, CR-ILD suggested with the experiments on BERT can boost the performance of T5 model as well as the encoder-based model. 
% Even though it has additional decoder architecture, unlike BERT, it shows similar tendencies that (1) distilling last layer knowledge is better than transferring all intermediate layers, (2) distilling only attention outperforms distilling attention and hidden states, and (3) applying CR-ILD can boost the model performance. 

\subsection{Experimental setup}
We experiment with our proposed training strategies on the encoder-decoder model.
As a teacher model, we use T5$_{\text{BASE}}$ fine-tuned to the target task with batch size 8, learning rate $1 \times 10^{-3}$ for ten epochs, which follows a training scheme for fine-tuning T5 on an individual GLUE task proposed in\,\cite{raffel2019exploring}. 
As a student model, we use the pre-trained T5$_{\text{Small}}$.
% The training scheme described above is also used during a distillation period. 
During the distillation, we distill the knowledge from the teacher model to the student model consecutively, similar to the training scheme described in the experimental setup of BERT distillation.
We first distill the knowledge using the given distillation objective (i.e., attention, intermediate states) depending on the task.
Unlike the BERT experiments, we fine-tune the T5 model on the target task after the ILD since the performance decreases in a few tasks when we apply logit-based KD\,\cite{hinton2015distilling}.
To distill the transformer layers and the intermediate states, we use methods proposed by\,\cite{wang2020minilmv2} and\,\cite{jiao2019tinybert}. 
Specifically, before distilling the attention scores, we applied relation heads proposed in \,\cite{wang2020minilmv2} and calculated attention scores since the number of attention heads of the student and the teacher differs.
After matching the number of relation heads, we distill attention scores of relation head and the hidden states, using the methods of \,\cite{jiao2019tinybert}.
Regarding the supplementary tasks, we use the same hyperparameters as the \texttt{ILD} experiments.
In \texttt{CR-ILD} experiments, we set $w^{\texttt{CR}}_{MHA}$ as 0.2 and 0.3 for the MRPC and RTE task individually.

\subsection{Experimental Results: Last Transformer Layer and Supplementary Task}
In this section, we focus on whether two findings from the experiments on BERT show consistent results in the experiments on T5.
% the effectiveness of compressing ED LMs with ILD, using proposed ILD methods. 
% For experiments, we apply T5$_{\text{BASE}}$ as a teacher and T5$_{\text{Small}}$ as a student on the GLUE benchmark. 
% We cannot directly apply MHA distillation since the number of attention heads between T5$_{\text{BASE}}$ and T5$_{\text{Small}}$ are different. Instead, we apply query-key relation distillation introduced in \,\citet{wang2020minilmv2}. 

\paragraph{Last Transformer Layer.} We evaluate the superiority of distilling the last Transformer layer knowledge in T5 models. 
Unlike BERT, T5 has an additional Transformer layer of the decoder network and cross-attention\,(CA). 
% 작성 내용: decoder vs encoder-decoder/ last vs uniform / ca include or not
Therefore, we also conduct additional comparisons between the distillation on the decoder network and the distillation on both the encoder and decoder network, as well as the comparison between the last Transformer layer mapping and uniform layer mapping.
Furthermore, we examine the effectiveness of the distillation on the cross-attention when we distill the knowledge in the decoder network.
% We further conduct additional comparisons for considering distilling this additional knowledge.

% decoder vs enc-dec 
In \autoref{fig:appx_t5_a}, \ref{fig:appx_t5_b}, and \ref{fig:appx_t5_c}, the blue boxes, and the orange boxes denote the distillation on the decoder network, the distillation on both the encoder and decoder network, respectively. 
 In most cases, distilling only from the decoder network tends to show higher results than distilling from the encoder and decoder network. 
% last vs unif
In addition, distilling the last Transformer layer shows better performance than the distilling Transformer layers uniformly.
% CA vs Self
Lastly, compared to distilling the self-attention and the cross-attention of the last Transformer decoder layer (green bar in \autoref{fig:appx_t5}), distilling only the self-attention of the last Transformer decoder layer (the first blue bar) shows better performance. 
In conclusion, We observe that distilling knowledge from only the last layer of the decoder network shows the highest performance across the target tasks. 
This result is consistent with the previous results of the experiments on BERT.
% By using CA distillation, we observe that distilling CA does not have advantages over distilling only self-attention (SA). 
% We further observe that using IR would lower the performance, and using only SA from the last Transformer layer of the decoder is the best choice. In most experiments, distilling CA shows lower performance than distilling only SA. 
% Furthermore, similar to BERT, we also observe that ILD for all Transformer layer incurs over-fitting to training data of RTE task in \autoref{fig:fig10d}.

\paragraph{Supplementary Task} We further evaluate the effectiveness of the supplementary tasks on ILD for the encoder-decoder models.
\autoref{fig:appx_t5_d} and \ref{fig:appx_t5_e} summarize the performance of RTE and MRPC tasks, dependiong on the supplementary task initialization.
Blue, red and orange lines denote distilling self-attention of the last Transformer layer, logit-based distillation, and fine-tuning, respectively.
Using the distillation on the self-attention of the last Transformer layer, initialization from the supplementary task training shows better performance than PLM initialization regardless of the supplementary task.
% We observe that using supplementary tasks with ILD can boost the performance of the encoder-decoder model . 
% \autoref{fig:appx_t5_d}, \ref{fig:appx_t5_e} summarize that the T5$_{\text{Small}}$ trained with ILD on other ST have higher performances for target task of MRPC and RTE, respectively. 
% Additionally, the brown bar in \autoref{fig:appx_t5_a}, \ref{fig:appx_t5_b}, and \ref{fig:appx_t5_c} show that applying \texttt{CR-ILD} is also effective in T5 models, which is consistent with results in BERT.

\subsection{Experimental Results: CR-ILD}
In this section, we examine whether our CR-ILD method could mitigate the over-fitting of the student model when the teacher and the student are T5 models.
In \autoref{fig:appx_t5_a}, \ref{fig:appx_t5_b}, and \ref{fig:appx_t5_c}, the brown box denotes to distill the self attention of last Transformer decoder layer with the consistency regularization, CR-ILD.
In order to see the difference according to the presence or absence of the consistency regularization, we compare the brown box and the first blue box, which denotes to distill the self attention of last Transformer decoder layer without CR-ILD. 
In the all tasks (MRPC, RTE, and SST-2), the consistency regularization boost the performance of the student model.
That is, the effect of the consistency regularization is consistent with the result of the experiment on BERT.
% 새로 바뀐 구성상 여기가 메인인데 이부분 분량이 좀 모자라네요 ...

% \paragraph{Loss landscape}

\end{document}